\documentclass{informs4}
\usepackage{comment}
\usepackage[utf8]{inputenc}
\usepackage{newunicodechar}
\newunicodechar{，}{,}

\usepackage{bbm}
\usepackage{algorithm}
\usepackage{algpseudocode}
\usepackage{tikz}
\usepackage{booktabs}
\usepackage{array}
\usepackage{tabularx}
\usepackage{graphicx}
\usepackage{subfigure}
\usepackage{multirow} 
\usepackage{natbib}
 \bibpunct[, ]{(}{)}{,}{a}{}{,}%
 \def\bibfont{\small}%
\EquationsNumberedThrough    % Default: (1), (2), ...
\TheoremsNumberedThrough     % Preferred (Theorem 1, Lemma 1, Theorem 2)
\ECRepeatTheorems  %  

\MANUSCRIPTNO{TRSC-0001-2024.00}

\begin{document}
%%%%%%%%%%%%%%%%

%\setcounter{page}{1}
%\DOI{10.1287/xxxx.0000.0000}%


% % Double Anonymous Review submission
% \RUNTITLE{Deep Reinforcement Learning Approach for DTSA-CCP}
% \TITLE{Improving After-sales Service: Deep Reinforcement Learning for Dynamic Time Slot Assignment with Commitments and Customer Preferences}

% % Single Anonymous Review submission
% \RUNAUTHOR{Xiao, Albert, Guohua, and Willem}

% % Title or shortened title suitable for running heads.
% % Enter the (shortened) title:
% \RUNTITLE{Deep Reinforcement Learning Approach for DTSA-CCP}

% \TITLE{Improving After-sales Service: Deep Reinforcement Learning for Solving Dynamic Allocation and Routing Problem with Preferred Time Windows}
\TITLE{Improving After-sales Service: Deep Reinforcement Learning for Dynamic Time Slot Assignment with Commitments and Customer Preferences}

\ARTICLEAUTHORS{%

\AUTHOR{Xiao Mao}
\AFF{School of Automation, Central South University, School of Industrial Engineering, Eindhoven University of Technology, Shenzhen Branch of China United Network Communications Co., Ltd.,
\EMAIL{maoxiao309@163.com}}

\AUTHOR{Albert H. Schrotenboer}
\AFF{School of Industrial Engineering, Eindhoven University of Technology,
\EMAIL{a.h.schrotenboer@tue.nl}}

\AUTHOR{Guohua Wu}
\AFF{School of Automation, Central South University, 
\EMAIL{guohuawu@csu.edu.cn}}

\AUTHOR{Willem van Jaarsveld}
\AFF{School of Industrial Engineering, Eindhoven University of Technology,
\EMAIL{W.L.v.Jaarsveld@tue.nl}}
} % end of the block

\ABSTRACT{%
\textbf{Problem definition:}
For original equipment manufacturers (OEMs), high-tech maintenance is a strategic component in after-sales services, involving close coordination between customers and service engineers. Each customer suggests several time slots for their maintenance task, from which the OEM must select one. This decision needs to be made promptly to support customers' planning. At the end of each day, routes for service engineers are planned to fulfill the tasks scheduled for the following day. We study this hierarchical and sequential decision-making problem—the Dynamic Time Slot Assignment Problem with Commitments and Customer Preferences (DTSAP-CCP)—in this paper.
\textbf{Methodology/results:} Two distinct approaches are proposed: 1) an attention-based deep reinforcement learning with rollout execution (ADRL-RE) and 2) a scenario-based planning approach (SBP). The ADRL-RE combines a well-trained attention-based neural network with a rollout framework for online trajectory simulation. To support the training, we develop a neural heuristic solver that provides rapid route planning solutions, enabling efficient learning in complex combinatorial settings. The SBP approach samples several scenarios to guide the time slot assignment. Numerical experiments demonstrate the superiority of ADRL-RE and the stability of SBP compared to both rule-based and rollout-based approaches. Furthermore, the strong practicality of ADRL-RE is verified in a case study of after-sales service for large medical equipment.
\textbf{Implications:} This study provides OEMs with practical decision-support tools for dynamic maintenance scheduling, balancing customer preferences and operational efficiency. In particular, our ADRL-RE shows strong real-world potential, supporting timely and customer-aligned maintenance scheduling.
}%

% \FUNDING{This research was supported by [grant number, funding agency].}

%Supplemental Material:
%Data Ethics & Reproducibility Note:

%\KEYWORDS{Stochastic programming, Decision support,Uncertainty, Disaster response, Optimization}

%\HISTORY{Received: Month DD, YYYY; Accepted: Month DD, YYYY; Published Online: Month DD, YYYY}

\maketitle
%%%%%%%%%%%%%%%%%%%%%%%%%%%%%%%%%%%%%%%%%%%%%%%%%%%%%%%%%%%%%%%%%%%%%%

% Text of your paper here
\section{Introduction}
Maintenance is critical to ensuring the reliability of high-tech systems used in radiosurgery, heating, ventilation, air conditioning, and image-guided therapy. Increasingly, original equipment manufacturers (OEMs) perform such maintenance on behalf of their customers. These after-sales services have become a significant revenue source for OEMs, accounting for approximately 28\%, 35\%, and 43\% of total revenue at Philips, Atlas Copco, and Elekta, respectively \citep{philips2024,atlascopco2023,elekta2025q3}. To make these services profitable, OEMs must carefully plan the deployment of scarce and specialized resources—particularly field service engineers (FSEs) who travel to geographically dispersed customer sites.

A key challenge in planning high-tech maintenance lies in synchronizing the availability of both the FSEs and the system equipment that has to be maintained. These systems are typically highly utilized and play a central role in customer operations. Thus, maintenance cannot be scheduled last-minute. Instead, customers must be informed of the maintenance moment well in advance to incorporate it into their planning. For instance, hospitals need to know when an image-guided therapy system will be offline for maintenance so they can adjust surgical schedules accordingly. Effective coordination between OEMs and their customers is, therefore, essential.

Motivated by these practices, we study a setting in which the system operator (i.e., the customer) and the OEM engage in a limited form of coordination, typically via telephone, whenever maintenance is upcoming. The operator first suggests a few convenient time slots, after which the OEM selects one that aligns with the availability of a qualified FSE who can travel to the site and perform the maintenance. The agreed time slot becomes a fixed point around which the operator structures their operational planning, regardless of whether it reflects their original preferences. We have learned that this approach is common in the high-tech business-to-business setting that we study \citep[see, e.g.,][]{van2024real}, as the telephone interaction also allows the sharing of contextual information that may help the OEM to better prepare for the visit. Comparable approaches are also adopted in consumer contexts, for example, when making an appointment for servicing a central heating unit \cite[see, e.g.,][]{ulmer2019enough}. 

Selecting maintenance service time slots in this setting is a complex real-time decision-making problem. The OEM must account for customer preferences, the availability and routing of FSEs, and the stochastic nature of future maintenance service requests. A key challenge—and a central novelty of our setting—is that the service provider selects and commits to a time slot immediately when a maintenance service request arrives, providing the customer with early certainty that is crucial for integrating the appointment into their own operational planning. This introduces a sequential structure in which the OEM must commit to time slots before knowing which other tasks will arise, and thus before any meaningful routes of the FSEs can be constructed. This structurally differs from the prior literature on dynamic multi-period routing problems \cite[see, e.g.][]{baty2024combinatorial}, where in each period it can be flexibly determined whether or not requests are postponed to a next period or serviced.

To address this challenge, we formulate and study a problem that captures these characteristics: the dynamic time slot assignment problem with commitments and customer preferences (DTSAP-CCP). It considers the real-time allocation of stochastically arriving customers (and their associated maintenance service requests) to future maintenance time slots and days. On each day, FSEs have to be routed along the maintenance service requests assigned to that day.  %under the assumption that the timing, quantity, and characteristics of customer maintenance requests are unknown in advance and revealed only upon arrival. Problems of this nature are typically modeled as multi-stage stochastic decision processes \citep{soeffker2022stochastic, baty2024combinatorial}. 
To reflect the balance between customer preferences and cost-efficiency, the DTSAP-CCP incorporates customer time slot preferences. If a customer is allocated to a time slot outside their preference, a preference penalty is incurred. Similarly, if on the day of execution the FSEs violate the time slot, a delay penalty is incurred. Besides these penalties, travel costs of the FSEs are incurred. The goal of the DTSAP-CCP is to minimize the expected total cost over a finite time horizon.

High-quality time slot assignments require incorporating long-term impacts into current decision-making. To do so, this paper proposes attention-based deep reinforcement learning with rollout execution (ADRL-RE). The attention-based deep reinforcement learning (ADRL) portion of our ADRL-RE is responsible for proposing real-time time slot assignments. Unlike existing methods that often rely on handcrafted rules or value approximations \citep{fleckenstein2023recent}, ADRL leverages an attention mechanism integrated with a gated recurrent unit (GRU), enabling the model to selectively focus on relevant temporal and contextual features during the sequential decision-making process. We then boost the ADRL performance with a rollout execution, yielding the ADRL-RE method as proposed in this paper.

We propose several key innovations within our ADRL-RE approach in order to combine rollouts with attention-based learning in a computationally efficient way. One should note that the end-of-day route planning for FSEs is essentially a vehicle routing problem with soft time windows (VRPSTW) - which is a complex combinatorial optimization problem. For actual end-of-day FSE route planning, we cannot rely solely on established heuristic solvers, such as OR tools \citep{ortools}, for solving the VRPSTW, as these heuristics are too slow for the learning phase of ADRL and the rollout execution, which requires solving the VRPSTW multiple times. To still obtain a method that has reasonable computational demands during training and a fast execution time, we propose a newly developed neural heuristic solver called the modified attention model (MAM) based on the attention model of \citet{koolattention}. We can then use MAM for both rapid rollout and efficient training of the ADRL model. This enables solving the VRPSTW in milliseconds, allowing rollouts to be executed in real-time, where decisions must be made in mere seconds. Effectively, we show that a neural heuristic solver can be used to speed up rollout execution and training of ADRL, making rollouts in complex combinatorial settings computationally feasible.

%The complexity  rollout method would incur significant online computational overhead, because while simulating decision alternatives via rollouts, expensive vehicle routing problems with soft time win our ADRL-RE method overcomes this by relying on two key innovations. First,  This eliminates the need for running computationally expensive heuristics in the rollout simulations.
%This method efficiently combines an online rollout framework, where each candidate time-slot assignment is simulated downstream, with an offline attention-based deep reinforcement learning (ADRL) method for time-slot assignment.
%To evaluate a rollout in mere seconds,  MAM effectively estimates the cost of solving a vehicle routing problem with soft time windows. Secondly, ADRL 
%Due to repeated simulations for each possible assignment, the rollout framework incurs significant computational overhead. Thus, a key innovation of ADRL-RE is its ability to execute rollouts in mere seconds. This is achieved by both the efficient time slot assignment of ADRL and a neural heuristic solver—modified from the attention model \citep{koolattention}, referred to as MAM—that estimates routing costs for FSEs, allowing for batch evaluation of routing problems.
% During inference, a high-performance heuristic solver is used to generate final routing plans. Together, ADRL and MAM for time slot assignment within the rollout framework, combined with OR-Tools for FSE routing, form our complete ADRL-RE.

The results show that the ADRL-RE outperforms several benchmark algorithms. One of these benchmarks is a newly developed scenario-based planning approach (SBP) based on the well-established scenario-sample framework \citep{hvattum2006solving}. SBP generates a set of sampled request scenarios and derives the current time slot assignment based on consensus across these scenarios. ADRL-RE outperforms both rule-based and rollout-based benchmarks, achieving significantly lower assignment penalties while maintaining modest travel costs. By means of a real-world case study based on medical institutions in Hunan Province, China, we illustrate the practical benefits of ADRL-RE.

The remainder of this paper is organized as follows. Section \ref{sec: LR} provides a review of the related literature. Section \ref{sec: PDM} introduces the DTSAP-CCP and formulates it as a Markov decision process (MDP). Section \ref{sec: SAS} details our ADRL-RE approach. Computational experiments and results are illustrated in Section \ref{sec: CE}. Finally, conclusions are drawn in Section \ref{sec: CON}.

\section{Literature Review} \label{sec: LR}

This paper contributes to two distinct streams of literature: the dynamic multi-period vehicle routing problem (DMPVRP) and the time window assignment vehicle routing problem (TWAVRP). In this section, we review the relevant literature from both streams and highlight our contributions. 

% From an application perspective, it is associated with dynamic VRPs across multiple periods, while from a methodological perspective, it relates to deep reinforcement learning (DRL). In this section, we review the relevant literature from both streams and highlight our contributions. 
% For a comprehensive overview of dynamic VRPs, we refer to \citet{zhang2023dynamic}, while for a general introduction to DRL, we refer to \citet{wang2022deep}. and \citet{hildebrandt2023opportunities} for reinforcement learning for dynamic VRP.

\subsection{Dynamic Multi-period Vehicle Routing Problem}
The DTSAP-CCP we introduce in our paper, with its emphasis on scheduling and planning after-sales activities, shares similarities with the DMPVRP; it also makes decisions from a range of combinatorial options at each decision point over multiple time periods. According to different applications, decision makers tackle the DMPVRP for selecting a delivery route, touting customers, restocking inventory, or dispatching technicians \citep{lagana2021dynamic, keskin2023dynamic, cuellar2024adaptive, jia2025scenario, chen2024technician}.

From a dynamic perspective, existing studies mainly focus on stochastic customer arrivals with temporal uncertainty. \citet{bonomi2024dynamically} investigate the DMPVRP with stochastic patients to optimize nurse-patient assignments and routing decisions. \citet{pham2024hybrid} introduce the DMPVRP with stochastic repair requests and demand. However, they typically defer decision-making until the end of each period, when a deterministic optimization problem is solved based on finalized customer sets. While this approach allows for information aggregation, it compromises response efficiency. To alleviate this issue, \citet{li2024reinforcement} propose a hybrid strategy that combines real-time dispatching with postponed decisions to address stochastic customer demands in parcel delivery systems. \citet{voccia2019same} further enhance this paradigm through third-party collaboration, enabling three operational options: immediate vehicle assignment, delayed assignment, or third-party outsourcing. Most recently, \citet{baty2024combinatorial} introduce temporal constraints by requiring stochastic customers to be serviced within predefined time windows, adding another layer of operational complexity.

While previous research has demonstrated operational efficacy in dynamic assignment and routing optimization, it has largely overlooked customer-centric aspects of service execution by not incorporating the customer perspective. Our study bridges this gap by integrating customer preferences into the real-time decision-making process, specifically taking into account each customer's preferred time windows. Furthermore, in the DTSAP-CCP, customers are assigned to a time slot for their service upon arrival at the system, whereas existing models and methods can decide to service customers at the start of a period. Although the latter option offers more flexibility for the decision maker, this is infeasible in the context of after-sales services in the high-tech industry.

\subsection{Time Window Assignment Vehicle Routing Problem}
By integrating the time window assignment with vehicle routing, TWAVRP determines delivery time slots and service routes to ensure satisfactory customer service \citep{agatz2011time}. Given the uncertainty in customer location, customer demand, and travel time, researchers develop methods for assigning time windows and planning visiting routes \citep{spliet2015time, yu2023time, ccelik2025exact}.

Since exact time window assignment for TWAVRP is computationally prohibitive for realistic problem sizes \citep{neves2018time}, most of the literature in this area focuses on designing heuristic solutions. In this context, \citet{campbell2005decision} propose a sampling-based heuristic insertion approach for managing time windows with stochastic customers in home delivery. \citet{kohler2020flexible} further develop this insertion approach by introducing the customer acceptance mechanisms that enable flexible time window management. \citet{bruck2018practical} address a related problem involving stochastic demand by developing a large neighborhood search heuristic for time slot management. While these heuristic approaches show operational feasibility in time window management, their application to large-scale TWAVRP remains severely limited by the prohibitive computational costs associated with determining time windows. This challenge arises from evaluating multi-scenario customer insertion operations across available time windows to guide the time window assignment. 

Recent advancements leverage machine learning to enhance time window assignment. \citet{van2024machine} propose a machine learning framework for predicting the feasibility of time slots for serving customers, significantly reducing the time-consuming process of feasibility checks. They leverage supervised learning to train a neural network, with training data generated using a commercial vehicle routing solver to solve deterministic VRPTW instances. Such supervised paradigms rely on a pre-optimized solution to obtain a label, whereas our study utilizes DRL to learn decision policies via interactions with the environment. More specifically, our ADRL-RE approach provides several innovations with regard to machine learning approaches in time window assignment. First, one of the building blocks of our approach is a neural heuristic solver of the VRPSTW, which we build upon the model provided by \cite{koolattention}. We then use this to enable quick evaluations of the VRPSTW both during training of the attention-based DRL model for time window assignment and while evaluating future trajectories during rollouts. We would like to emphasize that the ADRL, on its own, is the first application of attention models \citep{vaswani2017attention} for creating a time-window assignment policy in the context of dynamic multi-period vehicle routing.

\section{Problem Description and Modeling} \label{sec: PDM}
In this section, we first present a general overview of the DTSAP-CCP, followed by an illustrative example of its dynamics. Following this, we model the problem as a stochastic sequential decision problem and formulate it as an MDP.

\subsection{Problem Narrative}
The DTSAP-CCP considers an OEM utilizing a vehicle fleet $\mathcal{N}^v = \{1, \ldots, n_v\}$ to provide after-sales services over a time horizon $\mathcal{T} = \{0, 1, \cdots, T\}$, where each discrete time period $t \in \mathcal{T}$ corresponds to a single day. Each day, in turn, consists of a continuous time horizon of length $U$ during which service requests arrive at the system. All OEM's vehicles (with associated FSEs) originate from a single depot $0 = (x_0, y_0)$, where $x_0,y_0$ denote the coordinates of the depot. The service region is modeled as a graph $G = (\mathcal{N}, E)$, where the node set $\mathcal{N} = \{0\} \cup \mathcal{N}^c$ includes the depot and all  customer locations $\mathcal{N}^c = \{(x_1, y_1), \cdots, (x_{|\mathcal{N}^c|}, y_{|\mathcal{N}^c|})\}$, and $E$ represents the set of edges connecting these nodes. Each customer here represents a service request. Due to the stochastic and dynamic nature of service requests, the number of customers $|\mathcal{N}^c|$ is a stochastic variable.

In this context, the OEM's after-sales service process involves two departments. The first department is the customer service department, which is responsible for time slot assignment (TSA) by receiving service requests and assigning each customer a specific time slot. This happens in real-time between time $0$ and time $U$ within each day $t \in T$. Only time slots on day $t+1$ or later can be assigned. Each time slot $j$ has an earliest time $0 \leq a_j \leq U$ and a soft deadline $a_j \leq b_j \leq U$. We assume that time slot widths are fixed and correspond to a specific part of the day. Time slots can overlap, and multiple customer requests can be assigned to the same time slot. For example, there can be two time slots on each day, one in the morning between 8 am and 1 pm, and one in the afternoon between 12 pm and 17 pm. The second department is the maintenance department, which handles routing planning (RP) by organizing and deploying vehicles and technicians to customer locations in accordance with the assigned time slot. This happens at the end of each day $t$ for the routing on day $t+1$.

On day $t \in \mathcal{T}$, the TSA process starts upon the arrival of each customer $i \in \mathcal{N}^c$. The customer service team first gathers essential information, including the customer's location, and the customer's preferred time slots, to ensure a comprehensive understanding of the situation. We let the set $\mathcal I^t$ reflect all allocable time slots after day $t$, 
% and we define $\mathcal I$ as all time slots of the time horizon.
and the customer can be assigned to any time slot in $\mathcal{I}^t$. An assignment penalty $\alpha$ is incurred if the assigned slot does not align with the customer's $i$ preferences $C_i^p \subset \mathcal{I}^t$. This assignment is made directly when the customer arrives at the system. Once assigned and communicated, the time slot cannot be changed. Given that the number and details of customers remain unknown until they arise, the current TSA decision must account for potential future demands.  %This inherent uncertainty introduces significant complexity to the decision-making process, posing considerable challenges to making timely and effective decisions.

At the end of day $t$, the maintenance department executes the RP process for the customers scheduled for service on day $t+1$.  All vehicles depart from and return to the depot, with their routes fixed and unalterable once planned. The travel duration between nodes $i, j \in \mathcal{N}$ is determined by a deterministic travel time function $D(i, j)$, and each service has a duration $s$. Each customer $i \in \mathcal{N}^c$ must be visited exactly once. The actual RP then corresponds to specific arrival times at each customer at which the service starts. For time slot $j$, service cannot start before $a_j$, and a delay penalty of $\beta$ per time unit is incurred if service starts after $b_j$. Finally, travel costs equal the total duration of each vehicle route (excluding service time), i.e., the time spent traveling to and waiting at a customer location. The goal of the DTSAP-CCP is to minimize the expected sum of assignment penalties, delay penalties, and routing costs.

%Our DTSAP-CCP incorporates three types of costs: travel cost, delay penalty, and assignment penalty. The travel cost includes both the travel duration and any waiting time incurred when a vehicle arrives early. The delay penalty is imposed when services do not begin by the specified deadlines, while the assignment penalty depends on the time window decision. The objective of the DTSAP-CCP is to minimize the total costs over the entire time horizon. A simple example is provided in the following subsection to illustrate this process.

\subsection{Example of the DTSAP-CCP}
\begin{figure}
	\centering
	\includegraphics[width=1\linewidth]{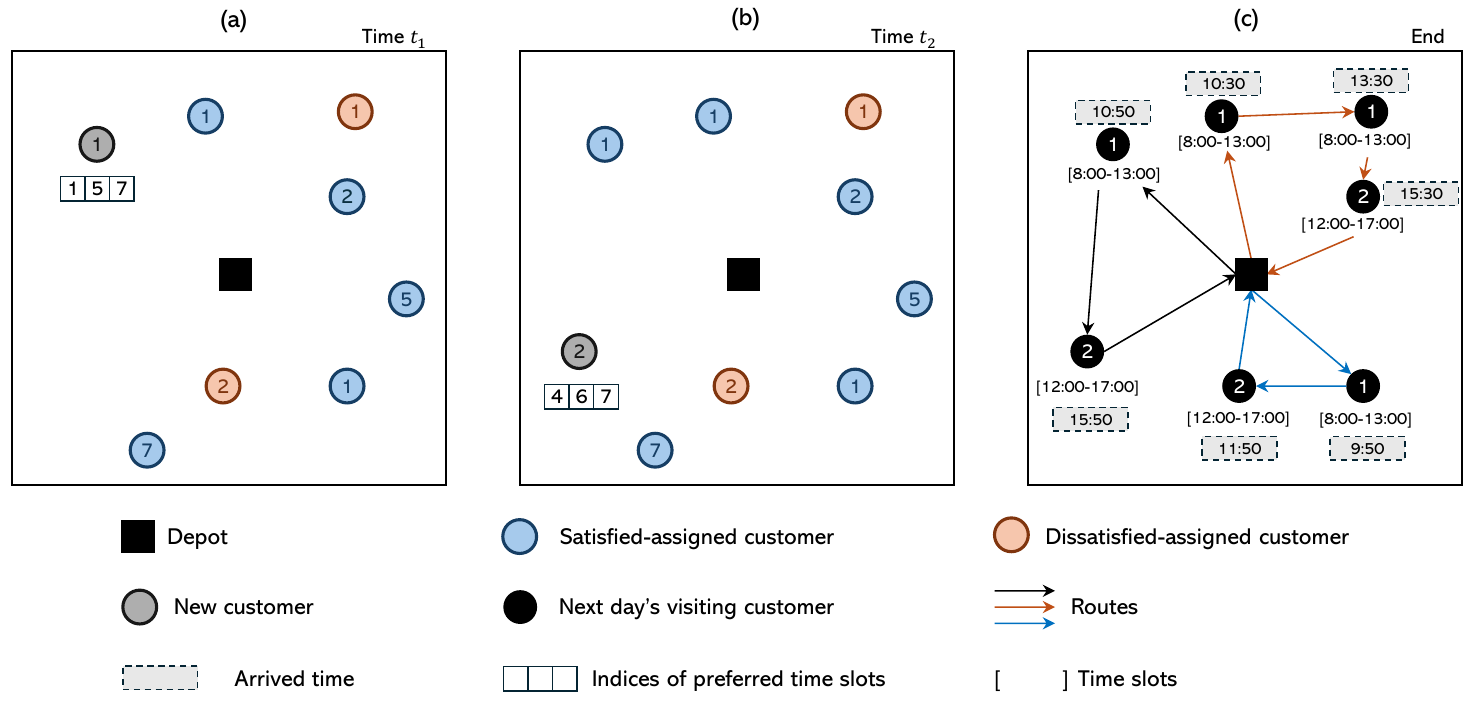}
	\caption{Illustrative example of DTSAP-CCP}
	\label{fig:2}
\end{figure}

Figure \ref{fig:2} illustrates two states of the TSA process and one state of the RP process for a given day.
% This example illustrates two states of the TSA process and the RP process for a given day, as shown in Figure \ref{fig:2}. 
Each customer has three preferred time slots and can be assigned to one of the time slots over the next five days. Each day is divided into two time slots: morning and afternoon. The numbers displayed on the customers indicate the indices of the assigned time slots. For example, a customer labeled with the number 5 corresponds to the morning of the third day following the current day.

At the time $t_1$ in Figure \ref{fig:2}(a), a new customer appears alongside seven previously assigned customers. Taking into account the locations, assigned time slots, and preferred time slots of all customers, the OEM determines a TSA decision that assigns time slot 1 (i.e., the morning of the following day) to the new customer, which aligns with the customer's preferences. Later, a similar situation arises in Figure \ref{fig:2}(b) with another new customer joining eight already assigned customers. This new customer is assigned to time slot 2, which falls outside the customer's preferences, thereby incurring an assignment penalty.

At the end of the day, the RP process is conducted to determine the visiting routes for customers scheduled for the following day, as illustrated in Figure \ref{fig:2}(c). Three vehicles, whose routes are represented by different colors, are deployed, with each customer requiring one hour of service. The morning and afternoon represent the time slots of 8:00 to 13:00 and 12:00 to 17:00, respectively, where their overlap is considered to enhance scheduling flexibility. Regarding the orange route, the second customer in the tour is visited after the soft deadline, incurring a delay penalty. In contrast, on the blue route, the vehicle arrives at its second customer before the earliest allowable time, necessitating a waiting period and resulting in waiting time costs.

\subsection{Markov Decision Process}
We now formalize the DTSAP-CCP as an MDP, detailing the decision epochs, state variable, decision variable, exogenous information, transition function, and objective function in the following subsections.

\subsubsection{Decision Epoch.}
The DTSAP-CCP concerns a finite discrete horizon $\mathcal{T}$ consisting of days $t = 0, 1, \cdots, T$. Each day, in turn, consists of a finite continuous time horizon of length $U$ that we model with a set of decision epochs $\mathcal{K}^t = \{1^t, \ldots, K^t, K^t + 1\}$. At each decision epoch, one of two events take place. If epoch $k^t \leq K^t$, a new customer arrives at the system, and a TSA decision must be made. If $k^t = K^t + 1$, the day has ended, and the vehicle fleet has to be routed to service the customers that were assigned a time slot on day $t + 1$. At $t=0$, a set of pre-existing customers $\mathcal{C}^{0}$ with pre-assigned time slots $\mathcal{M}^{0}$ is given. Note that $K^t$ is a stochastic variable whose realization is only known at the end of the day. 

\subsubsection{State Variable.} The state variable $S^t_k~=~(c^t_k, \mathcal C^t_k, \mathcal M^t_k)$ captures the information required to make decisions at decision epoch $k^t$. It consists of information regarding the new customer $c^t_k$, the customers $\mathcal{C}^t_k$ that have already a time slot assigned after day $t$, and the associated TSA decisions $\mathcal{M}^t_k$ of customers in $\mathcal C_k^t$. Here, $c^t_k$ contains $(x_i, y_i)$, the coordinates of the newly arrived customer, and $C_i^{p}$, the vector of $n_p$ preferred time slots of the newly arrived customer. Each element $c_i \in \mathcal C^t_k$ contains the corresponding customer coordinates $(x_i, y_i)$. Each customer's TSA decision $m_i \in \mathcal{M}^t_k$ is a time slot, so $m_i \in \mathcal{I}^t$. The relationship between the information of assigned customers and their TSA decisions is preserved by the ordering in the vectors $\mathcal{C}^t_k$ and $\mathcal{M}^t_k$, where the TSA decision associated with customer $c_i$ is $m_i$.

\subsubsection{Decision Variable.}
At each decision epoch $k^t$, the OEM has to assign a service time slot for the new customer (denoted by $x_k^{t, TSA}$) if $k^t \leq K^t$, or plan the visiting routes for the next day if $k^t = K^t + 1$ (denoted by $x_k^{t,RP}$). %To encode a decision, we define an indicator function $\mathbbm{1}_k$, which returns 1 if epoch $k$ is the last decision epoch of the day, i.e., $k = K+1$, and 0 otherwise. The value of $\mathbbm{1}_k$ indicates the decision type, where a TSA decision is made if $\mathbbm{1}_k = 0$, and a RP decision is made if $\mathbbm{1}_k = 1$. Thus, we define the space of decision variables at decision epoch $k$ by $X_k$, where each decision $x_k \in X_k$ is the combination of $x_k^{TSA}$, $x_k^{RP}$, and $\mathbbm{1}_k$, i.e., $x_k = (x_k^{TSA}, x_k^{RP}, \mathbbm{1}_k)$.
The decision $x_k^{t, TSA}$ is a time slot to which the customer $c^t_k$ is assigned. We assume that the customer can be assigned to any of the next $n_d$ days, with each day being divided into $n_s$ time slots. Therefore, we can write $x_k^{t,TSA} \in \mathcal{I}^t = \{(t - 1)n_s + 1, \dots, (t + n_d - 1)n_s\}$, and an assignment penalty $\alpha$ is incurred if $x_k^{t,TSA} \notin C_i^{p}$. %After the decision epoch $k = K$, all customer arrivals on day $t$ are assigned, and we turn to the next epoch for making decision $x_k^{RP}$.

If $k^t = K^t+1$, the decision $x_k^{t, RP}$ is a solution to the vehicle routing problem with soft time windows (VRPSTW), minimizing traveling and time window violation costs for the given TSA decisions on day $t+1$. Here, we define the $VRPSTW(G_t, x^t_k)$ as the objective value of the VRPSTW instance on graph $G_t$ when taking decision $x^t_k$. $G_t = (\mathcal{N}_t, E_t)$, where the node set $\mathcal{N}_t = \{0\} \cup \{\mathcal{N}_t^c\}$ includes the depot and customers $\mathcal{N}_t^c$ that is assigned to the day $t+1$, and $E_t$ represents the set of edges connecting these nodes. The customer set $\mathcal{N}_t^c$ is consists of the information $\mathcal{C}_t^*$ and TSA decisions $\mathcal{M}_t^*$ of the customers assigned to the next day. Each customer $i$ has a time slot $[a_t^i, b_t^i]$ according to its TSA decision, and we model the VRPSTW as a mixed-integer linear program. We define:  $y_t^{ijv}$ as a binary variable that equals 1 if vehicle $v \in \mathcal{N}^v$ traverses edge $(i, j) \in E_t$; $z_t^i$ as a continuous variable that denotes the arrival time at customer $i \in \mathcal{N}_t$; $w_t^i$ as a continuous variable that denotes the waiting time at customer $i \in \mathcal{N}_t$; and $d_t^i$ as a continuous variable that denotes the delay at customer $i \in \mathcal{N}_t$. Then, $VRPSTW(G_t, x^t_k)$ is defined as:
\begin{align}
    \text{min}& \sum_{v \in \mathcal{N}^v} \sum_{i \in \mathcal{N}_t} \sum_{j \in \mathcal{N}_t} D(i, j) y_t^{ijv} + \sum_{i \in \mathcal{N}_t^c} w_t^i + \beta \sum_{i \in \mathcal{N}_t^c} d_t^i, \label{eq1}\\
    \text{s.t.}& \sum_{v \in \mathcal{N}^v} \sum_{j \in \mathcal{N}_t} y_t^{ijv} = 1, && \forall i \in \mathcal{N}_t^c \label{eq2} \\
    & \sum_{i \in \mathcal{N}_t^c} y_t^{i0v} = \sum_{j \in \mathcal{N}_t^c} y_t^{0jv} \leq 1, && \forall v \in \mathcal{N}^v \label{eq3} \\
    & \sum_{j \in \mathcal{N}_t} y_t^{ijv} - \sum_{j \in \mathcal{N}_t} y_t^{jiv} = 0, && \forall v \in \mathcal{N}^v, \forall i \in \mathcal{N}_t^c \label{eq4} \\
    & z_t^j \geq z_t^i + s + D(i, j) + w_t^i - M(1 - y_t^{ijv}), && \forall v \in \mathcal{N}^v, \forall i, j \in \mathcal{N}_t, i \neq j \label{eq5} \\
    & z_t^i \geq a_t^i - w_t^i, && \forall i \in \mathcal{N}_t \label{eq6} \\
    & z_t^i \leq b_t^i + d_t^i, && \forall i \in \mathcal{N}_t \label{eq7} \\
    & z_t^0 = 0, \label{eq8} \\
    & y_t^{ijv} \in \{0, 1\}, && \forall v \in \mathcal{N}^v, \forall i, j \in \mathcal{N}_t, i \neq j \label{eq9} \\
    & z_t^i, w_t^i, d_t^i \geq 0. && \forall i \in \mathcal{N}_t \label{eq10}
\end{align}
%where $\beta$ is the penalty coefficient for delay. 
The objective function (\ref{eq1}) minimizes the total travel costs and delay penalties. Constraint (\ref{eq2}) ensures that each customer is visited exactly once, while constraint (\ref{eq3}) guarantees that each vehicle starts and ends at the depot. Notably, not all vehicles are required to visit customer nodes, which allows the model to optimize fleet utilization, thereby conserving resources and reducing operating costs. Constraint (\ref{eq4}) maintains flow conservation, and constraint (\ref{eq5}) enforces time continuity between consecutive nodes. Constraints (\ref{eq6}) and (\ref{eq7}) handle soft time windows, allowing early arrivals (with waiting) and late arrivals (with penalties). Constraint (\ref{eq8}) sets the departure time from the depot. Constraints (\ref{eq9}) and (\ref{eq10}) define the domain for the decision variables. 

Thus, the cost at decision epoch $k^t$ after taking decision $x^t_k$ in state $S^t_k$ is given by
\begin{equation} \label{objective function}
    C(S^t_k, x^t_k) = p^a(c^t_k, x^t_k) \mathbbm{1}_{k^t \leq K^t} + VRPSTW(G_t, x^t_k) \mathbbm{1}_{k^t = K^t+1},
\end{equation}
here, $p^a(c^t_k, x^t_k)$ equals $\alpha$ if $x_k^{t,TSA} \notin C_i^{p}$, and zero otherwise. 

% For the final decision epoch of day $t$, vehicle routes $x_k^{route}$ are constructed using $n^v$ vehicles to visit the customers allocated to the day $t + 1$. The travel duration between the customer and depot is given by a linear function $D$, calculated by multiplying a traveling coefficient with the distance, and each customer requires a service duration $d^s$. Determining the cost of route $r_k^v$ for vehicle $v$ involves calculating the traveling durations and evaluating each arrival time using the delay penalty function $P$. Specifically, $c_k^v = D(r_k^v) + \sum_{i \in r_k^v}P(l_i^v, tw_i^k)$, where $l_i^v$ denotes the time at which vehicle $v$ arrives at customer $i$ and $tw_i^k$ is the customer's time window. Thus, the cost at decision epoch $k$ after taking decision $x_k$ is given by $C(S_k, x_k) = p^a(c_k, x_k) (1 - \mathbbm{1}_k) + \sum_{v \in n^v}c_k^v \mathbbm{1}_k$ .

\subsubsection{Exogenous Information.}
The exogenous information after decision epoch $k^t$ evolves according to the following cases. When $k^t < K^t$, a new customer $c_{k+1}^t = (c_{k+1}^{t, r}, c_{k+1}^{t, l}, C_{k+1}^{t, p})$ arrives and is incorporated into the exogenous information $W_{k+1}^t$. When $k^t = K^t$, no new customer information is received and $c_{k+1}^t = \emptyset$ . Finally, when $k^t = K^t + 1$, the system advances to day $t+1$, and a new customer $c_1^{t+1} = (c_1^{t+1, r}, c_1^{t+1, l}, C_1^{t+1, p})$ is introduced into the exogenous information.

%which constructed from a newly appeared day ${c}_{t, k+1}^{*}$, a new location ${c}_{l, k+1}^{*}$, and a new vector of customer’s preferred time windows ${c}_{tw, k+1}^{*}$. In case $k+1$ is the last decision epoch of the day, the aforementioned set is empty.

\subsubsection{Transition Function.}
After executing action $x^t_k$, a transition is made from state $S^t_k$ to $S^t_{k+1}$. We split the transition into two parts: 1) the deterministic transition from state $S^t_k$ to a post-decision state $\hat{S}^t_k$, and 2) the stochastic transition from the post-decision state $\hat{S}^t_k$ to $S^t_{k+1}$, which depends on the realization of the exogenous information.

The transition towards the post-decision state $\hat{S}^t_k$ is associated with the action $x^t_k$, and distinct update steps are defined for different types of decisions. For $x_k^{t,TSA}$, a service time slot ${m}^t_k$ is assigned to the customer $c^t_k$, who then becomes an assigned customer. Thus, we define $\hat{S}^t_k~=~(\hat c^t_k, \hat{\mathcal{C}}^t_k, \hat{\mathcal M}^t_k)$ as follows:
\begin{itemize}
    \item $\hat{\mathcal{C}}^t_k := \mathcal{C}^t_k \cup c^t_k$: the set of assigned customers is updated to the union of the previous assigned customers and newly allocated customer.
    \item $\hat{\mathcal{M}}^t_k := \mathcal{M}^t_k \cup m^t_k$: the set of TSA decisions is updated to include the assigned time slot.
    \item $\hat{c}^t_k := \emptyset$ : the information associated with the new customer is cleared as its TSA decision is made.
\end{itemize}
For $x_k^{t,RP}$, visiting routes for customers assigned to the next day are planned based on their information $\mathcal{C}_t^*$ and TSA decisions $\mathcal{M}_t^*$. We update $\hat{S}^t_k$ in this type of decision as follows:
\begin{itemize}
    \item $\hat{\mathcal{C}}^t_k := \mathcal{C}^t_k \backslash \mathcal{C}_t^*$: the set of assigned customers is updated by removing customers scheduled to be visited the next day.
    \item $\hat{\mathcal{M}}^t_k := \mathcal{M}^t_k \backslash \mathcal{M}_t^*$: the set of TSA decisions is also updated by removing the TSA decisions of customers scheduled for the next day.
    \item $\hat{c}^t_k := \emptyset$ : assigning customer is still empty as no new customer arrives during the RP process.
\end{itemize}
The post-decision state is defined as $\hat{S}^t_k = \{\hat{c}^t_k, \hat{\mathcal{C}}^t_k, \hat{\mathcal{M}}^t_k\}$. Within a period, i.e., $k^t \leq K^t$ the state transitions to $S^t_{k+1} =\{c_{k+1}^t, \hat{\mathcal{C}}_k, \hat{\mathcal{M}}_k\}$. At the end of each period that $k^t = K^t +1$, the terminal state is used to initialize the next period via an inter-period transition $S^t_{k+1} = S^{t+1}_{1} = \{c_{1}^{t+1}, \hat{\mathcal{C}}_k, \hat{\mathcal{M}}_k\}$.

\subsubsection{Objective.}
The objective of the DTSAP-CCP is to find a decision policy $\pi \in \Pi$ that minimizes the total costs over the time horizon $\mathcal{T}$. The action $x^t_k$, determined under the state $S^t_k$, is guided by a decision rule $x^t_k = \chi^{\pi}(S^t_k)$. After the end of the time horizon, i.e., the end of day $T$, some customers are assigned to the next $n_d$ days, while only customers assigned to the following day are subject to RP decisions for visiting. Thus, we define a time period set $\mathcal{T}_o = \{T+2, \cdots, T+n_d\}$, where only RP decisions need to be made within these days for the unvisited customers. Thus, the objective function is written as follows:
\begin{equation}
    \mathop{\min}_{\pi \in \Pi} \enspace \mathbbm{E} \left[\mathop{\sum}_{t \in \mathcal{T}} \mathop{\sum}_{k \in \mathcal{K}^t} C(S^t_k, \chi^{\pi}(S^t_k)) + \mathop{\sum}_{t \in \mathcal{T}_o} C(S_1^t, \chi^{\pi}(S^t_1)) \right],
\end{equation}
where $S_1^t$ includes the information of unvisited customers on day $t \in \mathcal{T}_o$.
% \begin{equation}
%     \mathop{\min}_{\pi \in \Pi} \enspace \mathbbm{E} \left[\mathop{\sum}_{t \in \mathcal{T}} \mathop{\sum}_{k \in \mathcal{K}^t} C(S^t_k, \chi^{\pi}(S^t_k)) + \mathop{\sum}_{t \in \mathcal{T}_o} \mathop{\sum}_{k \in \mathcal{K}^t} VRPSTW(G_t, \chi^{\pi}(S^t_k)) \right],
% \end{equation}

\section{Solution Approaches} \label{sec: SAS}
We propose two approaches for solving the DTSAP-CCP. First, we propose an attention-based deep reinforcement learning with rollout execution (ADRL-RE). This method consists of several building blocks. First, it uses a neural heuristic solver to quickly obtain solutions to the end-of-day routing problem, i.e, the VRPSTW. This neural heuristic solver is called MAM and builds upon the attention model of \cite{koolattention}. Second, we propose an attention-based DRL (ADRL) model that provides a time-slot assignment policy. During training of the ADRL model, we use MAM to quickly solve the VRPSTW. Third, during its execution, the ADRL model is enhanced with a rollout mechanism that simulates future trajectories. Here, we also use the MAM model to quickly solve the VRPSTW in these future trajectories. Fourth, we employ OR-Tools for the VRPSTW instances that are actually being executed, i.e., those not being solved in training or during rollout: OR-Tools yields better solutions, but is not fast enough to be used during training or rollouts. We explain all these elements in Section 4.1.

Our second approach is a scenario-based planning approach (SBP), an online approach following a well-established framework in dynamic vehicle routing \citep{bent2004scenario, voccia2019same}. Here we do not use MAM as we need to solve multi-period VRPSTWs, and solely rely on OR-Tools. We explain all these elements in Section 4.2.

\subsection{Attention-based Deep Reinforcement Learning with Rollout Execution}\label{sec:4.1}

The ADRL-RE approach integrates an offline-trained attention-based deep reinforcement learning (ADRL) model with an online decision refinement mechanism implemented via a rollout framework. The following sections describe the neural network architecture, the offline training procedure, and the design of the rollout framework for online decision enhancement.

\subsubsection{Network Architecture of ADRL.}
The ADRL model is responsible for time slot assignment, which takes $S^t_k$, the state at decision epoch $k^t$, as input. For notational simplicity, we omit the day index $t$ in the following descriptions. To capture the necessary information about the depot, the new customer, and assigned customers, we construct the feature set for each customer $i$ with attributes: 1) coordinates $c_i^l$, representing the customer's location, 2) appearance day $t_i$, where the depot and pre-existing customers $\mathcal{C}^0$ are assigned a value of zero, 3) preferred time slots $C_i^p$, where the depot’s preferences is represented by a zero-valued set, and 4) the TSA decision $m_i$, indicating the time slot assignment, and the new customer is set with a value of zero. 

\begin{figure}
	\centering
	\includegraphics[width=1\linewidth]{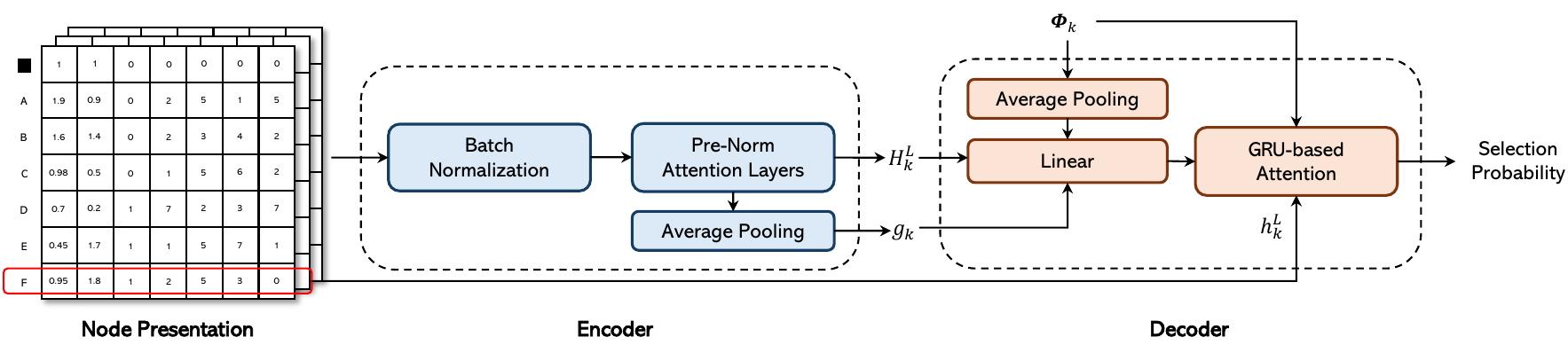}
	\caption{Network Architecture}
	\label{fig:3}
\end{figure}

Here, a Transformer-style network architecture is designed, featuring an encoder built with a slack of Pre-Norm attention layers \citep{xiong2020layer} and a decoder with a GRU-based attention layer. An overview of this ADRL model is illustrated in Figure \ref{fig:3}. At decision epoch $k$, the encoder processes the state by first applying batch normalization, followed by $L$ Pre-Norm attention layers, to generate the node embeddings $H_k^L$, where we denote $h_k^L$ as the node embedding of the new customer. Subsequently, an average pooling operation is applied across the node embeddings to produce the graph embedding $g_k$. This pooling operation aggregates the information from all nodes, capturing the overall context of the current state. 

At the beginning of decoding, we construct a TSA embedding $\Phi_k$ by aggregating and pooling the node embeddings of customers assigned to each time slot. Specifically, we define:
\begin{equation}
    \Phi_k = [\max(\Phi_k^1), \cdots, \max(\Phi_k^n)]
\end{equation}
where $n = n_d n_s$ is the total number of available time slots for assignment, and $\Phi_k^1$ represents the node embeddings of customers assigned to the first time slot of the following day. The decoder takes the TSA embedding $\Phi_k$, node embeddings $H_k^L$, and graph embedding $g_k$ as input. After applying an average pooling operation to $\Phi_k$, the result is concatenated with $H_k^L$ and $g_k$ to generate a hidden vector $v_k$ through a linear projection. The GRU-based attention layer then processes $\Phi_k$, $v_k$, and $h_k^L$ to generate the selection probability for each available action. This layer captures both the temporal dynamics and spatial relationships among customers and the depot by attending to the relevant parts of the state during decoding. Based on the generated probabilities, the assigned time slot can be determined using either greedy or sampling strategy. The greedy strategy selects the time slot with the highest probability, while the sampling strategy selects a time slot by sampling according to its probability. Details on the Pre-Norm attention and GRU-based attention layers can be found in Appendix A.

\subsubsection{Offline Training}
To obtain the neural network decision policy for TSA, we train our ADRL model in an offline simulation fashion. For each training iteration $i$, a training dataset $\mathcal{D}_i$ is generated based on distributions of arrival times, locations, and preferred time slots. A DTSAP-CCP instance $u \in \mathcal{D}_i$ consists of pre-existing customers on $t=0$ and new customers throughout the time horizon $\mathcal{T}$. On each day $t$, the ADRL model needs to make a total of $K^t$ TSA decisions. 

At the end of the day, a route solver is used to solve the VRPSTW to make the RP decision for the next day. Although heuristic routing solvers have been extensively studied, they often suffer from limited computational efficiency, primarily due to their lack of support for batch processing. This limitation poses a challenge for achieving stable and rapid convergence in the reinforcement learning model. 

DRL-based routing approaches, such as AM \citep{koolattention}, have demonstrated strong potential in solving vehicle routing problems. However, they often overlook the time windows of customers. To address this issue, we propose the MAM by integrating a Pre-Norm attention layer into the encoder and designing two decoders: one for vehicle selection and the other for customer selection. The details of MAM are provided in Appendix B. Based on MAM, the ADRL-MAM approach is developed to support the training of the ADRL model. For each new customer, the ADRL is run to determine the TSA decision, taking into account the customers’ preferred time slots. At the end of each day, the MAM is employed to plan the vehicle routes for visiting the customers scheduled for the following day. 

In addition, the uncertainty in the location, preferred time slots, and the number of customers arriving daily also makes the effective training of the ADRL model challenging. In a practical scenario, when a new customer arrives, both the encoder and decoder of the ADRL model must run once to determine the action. However, this "1-encode-1-decode" data processing mode introduces inefficiencies and instability in model training due to the time-consuming nature of the encoding process. To mitigate this, we propose a "1-encode-N-decode" mode for the offline training, where all customers arriving on the same day are encoded together and then decoded sequentially to determine the TSA decisions. In this way, training efficiency is improved by reducing the frequency of the computationally expensive encoding operations. Notably, the number of customers appearing on the same day varies across instances within a training batch, leading to inconsistent input lengths and hindering batch processing. We pad the data with depot features, ensuring same input lengths across all instances in the training batch. 

\begin{algorithm}[htbp]
    \begin{algorithmic}[1]
        \Require initial policy network parameter $\theta_0$, initial baseline network parameter $\theta_0^b$, number of training iterations $I$.
        \Ensure $\theta_1, \cdots, \theta_I$.
        \For {$i = 0, \cdots, I-1$}
            \State Generate training dataset $\mathcal{D}_i$ and construct batch instances $u$ from $\mathcal{D}_i$
            \For {$t = 1, \cdots, T$ in $u$}
                \For {$k = 1, \cdots, K^t$}
                    \State{$x_k^t \gets \mathrm{ADRL}_{\theta_i}(S_k^t)$ }
                    \State{Find exogenous information $c_{k+1}^t$}
                    \State{$S_{k+1}^t \gets$ $\text{Transition}(S_k^t, x_k^t, c_{k+1}^t)$}
                \EndFor
                \State{$x_{k+1}^t \gets \mathrm{MAM}(S_{k+1}^t, x_{k+1}^t)$}
                \State{Find exogenous information $c_1^{t+1}$}
                \State{$S_{1}^{t+1} \gets \text{Transition}(S_{k+1}^t, x_{k+1}^t, c_1^{t+1})$}
            \EndFor
            \State{Get the objective value of policy network $\mathcal{L}(\sigma)$ according to eq. (\ref{objective function})}
            \State{Get the objective value of baseline network $\mathcal{L}^b(u)$}
            \State{$\theta_{i+1},\theta_{i+1}^b \gets \mathrm{REINFORCE}(\mathcal{L}(\sigma), \mathcal{L}^b(u))$}
        \EndFor
    \end{algorithmic}
    \caption{"1-encode-N-decode" Training}
    \label{training_algo}
\end{algorithm}

We outline training process, which executes $I$ training iteration, in Algorithm \ref{training_algo}. After generating the training dataset (line 2), we use the ADRL model to make TSA decisions (line 5). At the end of each day, the pre-trained MAM is employed to plan the visiting routes of the next day (line 9). Following \citet{koolattention}, \citet{xu2021reinforcement}, and \citet{pan2023h}, the REINFORCE algorithm with a greedy rollout baseline is utilized to compute the loss gradient with respect to the network parameters $\theta$, as follows:
\begin{equation}
    \nabla\mathcal{M}(\theta) = \mathbb{E}_{p_{\theta}(\sigma | u)} \left[(\mathcal{L}(\sigma) - \mathcal{L}^b(u)) \nabla\mathrm{log}p_{\theta}(\sigma | u) \right],
\end{equation}
where $p_{\theta}(\sigma | u)$ is the probability distribution from which a solution (TSA decisions) $\sigma$ can be sampled, given the instance $u$. $\mathcal{L}(\sigma)$ denotes the objective value of the solution $\sigma$, while $\mathcal{L}^b(u)$ represents the objective value for instance $u$ solved by the greedy baseline network. During the training of the ADRL model, the sampling strategy is utilized, whereas greedy strategy is employed during inference.

\subsubsection{Rollout Framework} \label{sec: SA}
Although the previously presented end-to-end ADRL (EE-ADRL) offers a time-efficient way to directly obtain TSA decisions, the solution quality generated by end-to-end approaches cannot be guaranteed, especially for complex problems and network architectures \citep{glasmachers2017limits, li2022overview}. To address this issue, inspired by AlphaZero \citep{silver2018general}, we design a rollout framework based on the ADRL to enhance the search for better solutions. To solve the VRPSTW instances generated from the TSA decisions, we utilize heuristic routing solver OR-Tools \citep{ortools}. The entire approach is termed ADRL-RE, and its time slot assignment procedure based on the rollout framework is outlined in Algorithm \ref{rollout_framework}.

\begin{algorithm}[htbp]
    \begin{algorithmic}[1]
        \Require the current decision epoch $k^t$, the new customer $c_k^t$, the assigned customers $\mathcal{C}_k^t$, the TSA decisions $\mathcal{M}_k^t$, the quantity of rollouts $m$, the trained ADRL model, and the neural heuristic solver MAM.
        \Ensure the TSA decision for the new customer $x_k^t$.
        \State {$A \gets \text{ADRL}(c_k^t, \mathcal{C}_k^t, \mathcal{M}_k^t)$}
        \For {$i = 1, \cdots, m$}
            \State {Construct rollout customers $\widetilde{\mathcal{C}}_i$ from current decision epoch $k^t$ to the end of time horizon}
            \For {$a \in A$}
                \State $RD_i \gets \text{Generate\_rollout\_dataset}(c_k^t, a, \widetilde{\mathcal{C}}_i, \mathcal{C}_k^t, \mathcal{M}_k^t)$
                \State $O_a^i \gets \text{Calculate\_objective\_value}(RD_i, \text{ADRL}, \text{MAM})$
            \EndFor
	\EndFor
        \State $O_a \gets \frac{1}{m} \sum_{i=1}^{m} O_a^i, \forall a \in A$
        \State $x_k^t \gets \mathop{\arg\min}\limits_a(O_a)$
    \end{algorithmic}
    \caption{Time Slot Assignment Procedure Based on the Rollout Framework}
    \label{rollout_framework}
\end{algorithm}

In contrast to the end-to-end pattern, which determines the final decision using either a greedy or sampling strategy, the ADRL-RE constructs rollouts for promising actions $A$ to enhance the solution search capability. When a new customer $c_k^t$ arrives, the ADRL model is first utilized to generate the selection probabilities for the available actions. Then, the rollout process is performed in ADRL-RE for promising actions with non-zero selection probabilities to refine the final decision. The details of the rollout process are as follows:

\begin{itemize}
    \item \textbf{Rollout Customers Construction:} following the same distribution used for training data generation, we sample customers who are expected to arrive until the end of the time horizon as rollout customers, representing an approximation of the future. This process repeated $m$ times, resulting in $m$ sequences of rollout customers.
    \item \textbf{Rollout Dataset Generation:} by integrating data concerning appeared customers, the new customer, and rollout customers, we create a rollout instance. Therefore, for each promising action taken for the new customer, $m$ rollout instances are generated, resulting in a rollout dataset of $m \times pa$ instances, where $pa$ is the number of promising actions. Since rollout instances may have diffrent number of customers, we pad all rollout instances to the same length using the depot feature to facilitate batching processing.
    \item \textbf{Rollout Decision Selection:} the instances in the rollout dataset are addressed using ADRL-MAM to improve efficiency. After that, we compute the average objective value for each promising action across the corresponding $m$ rollout instances. This average objective value reflects the action quality, taking into account both the previous customers and potential future customers. The action with the lowest objective value is then selected as the final decision for the new customer.
\end{itemize}

\subsection{Scenario-based Planning Approach}
At each decision epoch where a TSA decision is required, the SBP anticipates future customers by sampling a set of scenarios according to the current state. For each sampled scenarios, a corresponding deterministic routing problem is generated. By solving it, the TSA decision for each scenario is derived, guiding the action selection. Given the extensive research on VRPs, a well-established and validated heuristic solver is utilized at the RP decision epoch to efficiently solve the VRPSTW. 

\subsubsection{Scenario-based Planning for TSA.}
On day $t$, at each decision epoch $k^t \in \{1,\cdots, K^t\}$, the SBP needs to make TSA decision for the newly arrived customer. To support this decision, we generate a set of scenarios $\mathcal{Q} = \{1, 2, \cdots, q\}$, where a set of future customers $Q_l$ is sampled in each scenario $l \in \mathcal{Q}$. As presented in Algorithm \ref{SBP}, our SBP determine the TSA decision via three key components. For each TSA decision epoch $k^t$, the SBP takes information of the assigned customers $\mathcal{C}_k^t$ and their TSA decisions $\mathcal{M}_k^t$, the current customer $c_k^t$, the number of sampled scenarios $q$, and parameters related to the arrival and preferred time slots distributions $P_{dis}$ as input.

\begin{algorithm}[htbp]
    \begin{algorithmic}[1]
        \Require $\mathcal{C}_k$, $\mathcal{M}_k^t$, $c_k^t$, $q$, $P_{dis}$.
        \Ensure $x_k^{t, TSA}$.
        \State $\mathcal{Q}_k^t \gets \mathcal{F}_{gen}(\mathcal{C}_k, \mathcal{M}_k^t, c_k^t, q, P_{dis})$
        \State $\psi_k^t \gets \mathcal{F}_{plan}(\mathcal{Q}_k^t)$
        \State $x_k^{t, TSA} \gets \mathcal{F}_{dec}(\psi_k^t)$
    \end{algorithmic}
    \caption{Scenario-based Planning for TSA}
    \label{SBP}
\end{algorithm}

The first component is the function $\mathcal{F}_{gen}(\mathcal{C}_k, \mathcal{M}_k^t, c_k^t, q, p_{dis})$, which is responsible for generating sampling scenarios. Each scenario $l$ includes both the assigned customers $\mathcal{C}_k^t$, the current customer $c_k^t$ requiring the time slot assignment, as well as potential future customers $Q_l$ generated from a probability distribution. The parameter set $P_{dis}$ contains all necessary distribution parameters for generating the number, locations, and preferred time slots of the future customers. A key element in $P_{dis}$ is the sampling horizon, which specifies how far into the future customers should be generated. From the results of systematic experimental tests in Appendix C, scenarios for the remaining customers of the current day are sampled.

The next step in Algorithm \ref{SBP} is to generate scenario plans, achieved by the function $\mathcal{F}_{plan}(\mathcal{Q}_k^t)$. It aims to solve the corresponding deterministic multi-period VRPSTW for each sampled scenario $l \in \mathcal{Q}_k^t$, where a graph $G_l^{t,k}=(\mathcal{N}_l^{t,k}, E_l^{t,k})$ is defined on time horizon $\mathcal{T}_k^t$. The node set $\mathcal{N}_l^{t,k} = \{0\} \cup \mathcal{C}_k^t \cup \{c_k^t\} \cup Q_l$ comprises the depot, assigned customers $\mathcal{C}_k^t$, the current customer $c_k^t$, and sampled future customers $Q_l$, while $E_l^{t,k}$ represents the set of edges connecting these nodes. 

Each customer $i$ is associated with a set of time slots $\mathcal{W}_i$. For assigned customers, a specific time slot on their scheduled day is given by $\mathcal{W}_i = \{(t_j,[a_i, b_i])\}$, where $t_j \in \mathcal{T}_k^t$ denotes the scheduled day and $[a_j, b_j]$ represents the assigned service time window. In contrast, current and future customer $i$ are associated with multiple preferred time slots spanning several days, denoted by $\mathcal{W}_i = \{(t_{j1},[a_{i1}, b_{i1}]), (t_{j2},[a_{i1}, b_{i1}]), \cdots \}$, where each element corresponds to a preferred time slot and its corresponding period. Since the actual service periods and time windows of current and future customers are not yet determined during scenario planning, their visits are restricted to the days corresponding to their preferred time slot. Therefore, the set of time slots for the current customer is $\mathcal{W}_i = \{(t_{j1},[0, t_{end}]), (t_{j2},[0, t_{end}]), \cdots \}$, where $t_{end}$ is the end time of each day. For future customers, we assumed sufficiently wide service windows on their preferred days, represented by $\mathcal{W}_i = \{(t_{j1},[0, \infty]), (t_{j2},[0, \infty]), \cdots \}$. This relaxed formulation promotes more efficient routing by avoiding overly restrictive temporal constraints, as further discussed in Appendix C.

The final step in Algorithm 1 is to determine the time slot assignment for the current customer via the function $\mathcal{F}_{dec}(\psi_k^t)$. Each scenario plan $sp_i \in \psi_k^t$ provides information about the visit time for the current customer, which allows us to obtain the time window assigned to the customer. The decision $x_k^{t, TSA}$ is then determined using a frequency-based consensus mechanism, where the assigned time slot corresponds to the option most frequently selected across all scenario plans. 
% This selection strategy is both computationally efficient and robust, as it filters out temporary scenario-specific variations and retains the most common temporal pattern from multiple scenarios.

\subsubsection{Determining Routing Solution by OR-Tools.}
During the RP process at decision epoch $k^t=K^t+1$, the customers scheduled to receive service the next day are identified. The core decision-making task involves solving the derived VRPSTW to determine the route plan $x_k^{t, RP}$. While numerous solvers have been developed for VRPs \citep{pessoa2020generic, erdougan2017open, Wouda_Lan_Kool_PyVRP_2024}, most do not support soft time windows due to the extensive search space. In this study, we apply OR-Tools to solve this VRPSTW and obtain visiting routes. Once these routes are established, they remain fixed during execution. OR-Tools provides an efficient means of obtaining near-optimal solutions for VRPSTW within a reasonable runtime. However, it is important to highlight that our SBP framework is designed to be adaptable, allowing for the integration of alternative routing methods as needed to suit specific requirements.

\section{Computational Experiments} \label{sec: CE}
In this section, we first present the setup of the experiments, followed by a detailed description of the ADRL training procedure. Then we provide results of our ADRL-RE and SBP approaches compared to various benchmark algorithms. All experiments are conducted on two nodes of a computer cluster. Model training and testing are performed on a GPU node, while other approaches are executed on a CPU node. The GPU node is equipped with two Intel Xeon Platinum 8360Y processors, 120 GB of RAM, 16 CPU cores, and an NVIDIA A100 GPU with 40 GB of memory. The CPU node features AMD Rome 7H12 processors, 56 GB of RAM, and 32 CPU cores. All code is developed in Python 3.9 with PyTorch 2.1.

\subsection{Experiment Configuration}
Our experiments are conducted on a 10-day time horizon, where each day consists of nine basic working hours, divided into a morning time slot 8:00-13:00 and an afternoon time slot 12:00-17:00. The TSA decision for each customer can be any time slot within the next five days. We set the penalty cost $\alpha = 2$ and the delay penalty $\beta = 3$. To comprehensively evaluate the performance of our approaches, we construct six systems that vary in the number of customers and vehicles.

\subsubsection{Instance Generation}
Each 10-day simulation of our system is referred to as an instance. During the instance generation, two types of customers are considered: 1) pre-existing customer $\mathcal{C}^0$, which are pre-existing on day 0, and 2) daily customer $c_k^t$, which appear dynamically each day throughout the time horizon. The coordinates of customers are randomly generated in a 2-unit square service area with the depot centrally located. The number of pre-existing customers is sampled from a normal distribution $N(n_{pre}, 3)$ with rounding. Since these pre-existing customers have already been assigned, we assume that their assignments satisfy their preferred time slot. For daily customers appearing on day $t$, the number is also sampled from a normal distribution $N(n_{daily}, 3)$ with rounding. Each daily customer $c_k^t$ is assigned with three preferred time slots, which are randomly sampled from the time slot set of next five days $\mathcal{I}^t$. Here, $\mathcal{I}^t = \{2t-1, \cdots, 2t+8\}$ denotes the 10 consecutive time slots within the next five days, ordered from the nearest to the farthest relative to $t$. For example, $2t-1$ corresponds to the morning of the next day, $2t$ to the afternoon of the next day, and so on until $2t+8$. Furthermore, we consider different vehicle fleet with 2, 3, and 4 vehicles for RP. The objective value in each instance is associated with two parameters $p_{tra}$ and $p_{ser}$, where the former defines the traveling coefficient and the latter defines the service duration at each customer. Six distinct systems are constructed for the experiments, with corresponding parameters illustrated in Table \ref{tab:1}.

\begin{table}[htbp] \footnotesize
  \centering
  \caption{System Parameters}
  \renewcommand\arraystretch{1.2}
    \begin{tabular}{cccccc}
    \toprule
    \makebox[0.1\textwidth][c]{System} & \makebox[0.08\textwidth][c]{$n_{pre}$} & \makebox[0.08\textwidth][c]{$n_{daily}$} & \makebox[0.08\textwidth][c]{$n_v$} & \makebox[0.06\textwidth][c]{$p_{tra}$} & \makebox[0.08\textwidth][c]{$p_{ser}$} \\
    \midrule
    S1& 30 & 15 & 2 & \makebox[0.04\textwidth][c]{\hfill 1.0} & \makebox[0.05\textwidth][c]{\hfill 0.6667} \\
    S2& 30 & 15 & 3 & \makebox[0.04\textwidth][c]{\hfill 1.5} & \makebox[0.05\textwidth][c]{\hfill 1.0}  \\
    S3& 30 & 15 & 4 & \makebox[0.04\textwidth][c]{\hfill 2.0} & \makebox[0.05\textwidth][c]{\hfill 1.3333}  \\
    
    S4& 40 & 20 & 2 & \makebox[0.04\textwidth][c]{\hfill 0.75}  & \makebox[0.05\textwidth][c]{\hfill 0.5}  \\
    S5& 40 & 20 & 3 & \makebox[0.04\textwidth][c]{\hfill 1.125} & \makebox[0.05\textwidth][c]{\hfill 0.75}  \\
    S6& 40 & 20 & 4 & \makebox[0.04\textwidth][c]{\hfill 1.5}   & \makebox[0.05\textwidth][c]{\hfill 1.0}  \\
    \bottomrule
    \end{tabular}%
  \label{tab:1}%
\end{table}%

\subsubsection{Parameter Settings}
For the ADRL network architecture, we implement the same hyper-parameters as reported in \citet{koolattention} and conduct 40 training iterations. In each iteration, $1,280,000$ instances are generated for training with a batch size of 512. The Adam optimizer is used to update the network parameters, with an initial learning rate of $10^{-4}$. To ensure stability during training, the norm of the network gradient is clipped within 5.0, and the learning rate decays with a parameter of 0.996. For determining the number of rollouts in ADRL-RE, we conduct an experiment in system S1 with 5, 10, and 20 rollout instances. The results in Appendix D indicate that ADRL-RE, when executed with 10 rollout instances, effectively balances the objective value, customer satisfaction, and time efficiency, demonstrating excellent overall performance. Therefore, we employ ADRL-RE with 10 rollout instances in subsequent experiments.

\begin{figure}[htbp]
	\centering
        \subfigure[S1]{
        \includegraphics[width=0.31\linewidth]{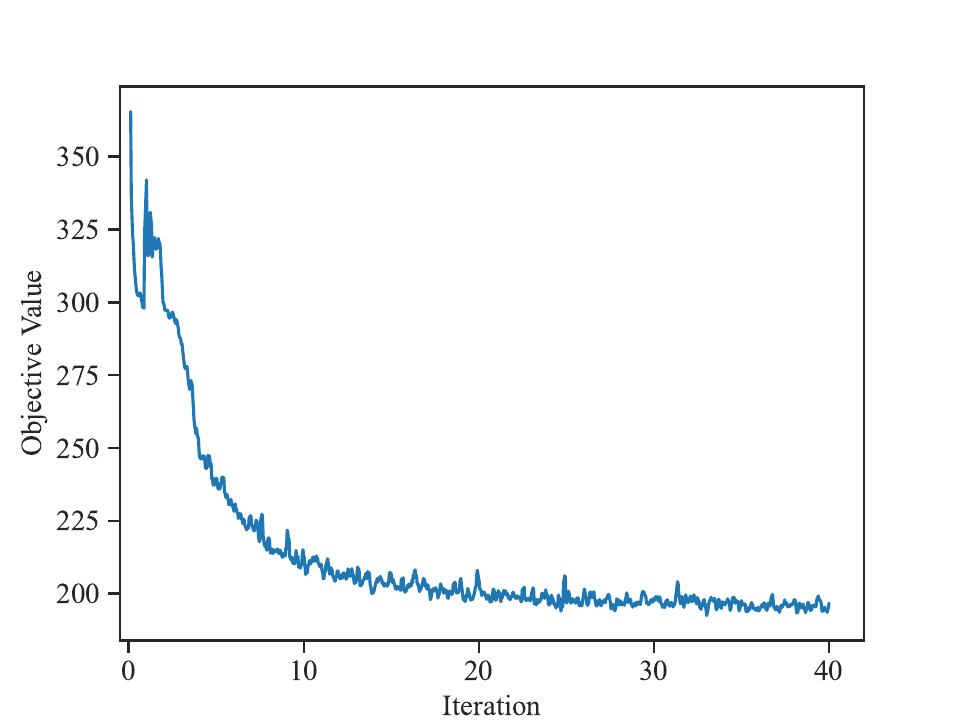}
        }
        \subfigure[S2]{
        \includegraphics[width=0.31\linewidth]{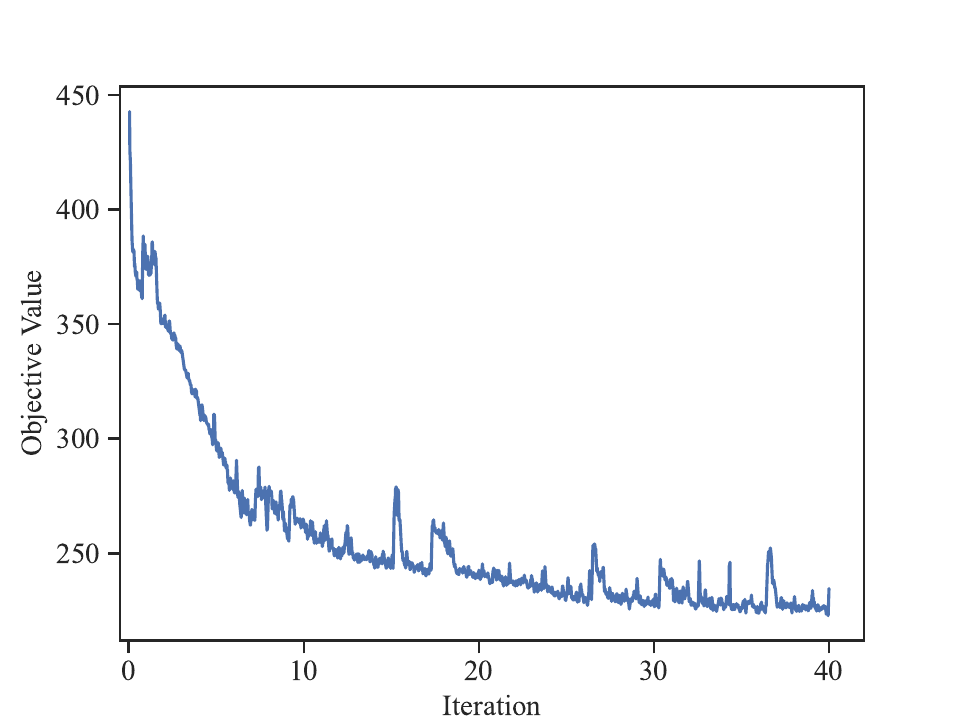}
        }
        \subfigure[S3]{
        \includegraphics[width=0.31\linewidth]{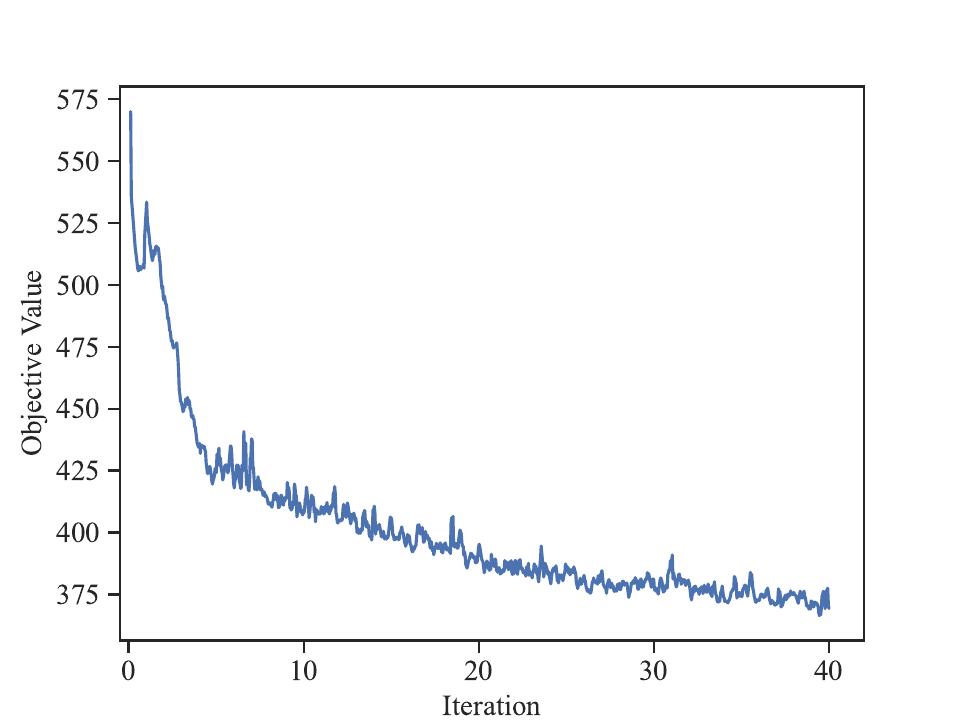}
        }
        \subfigure[S4]{
        \includegraphics[width=0.31\linewidth]{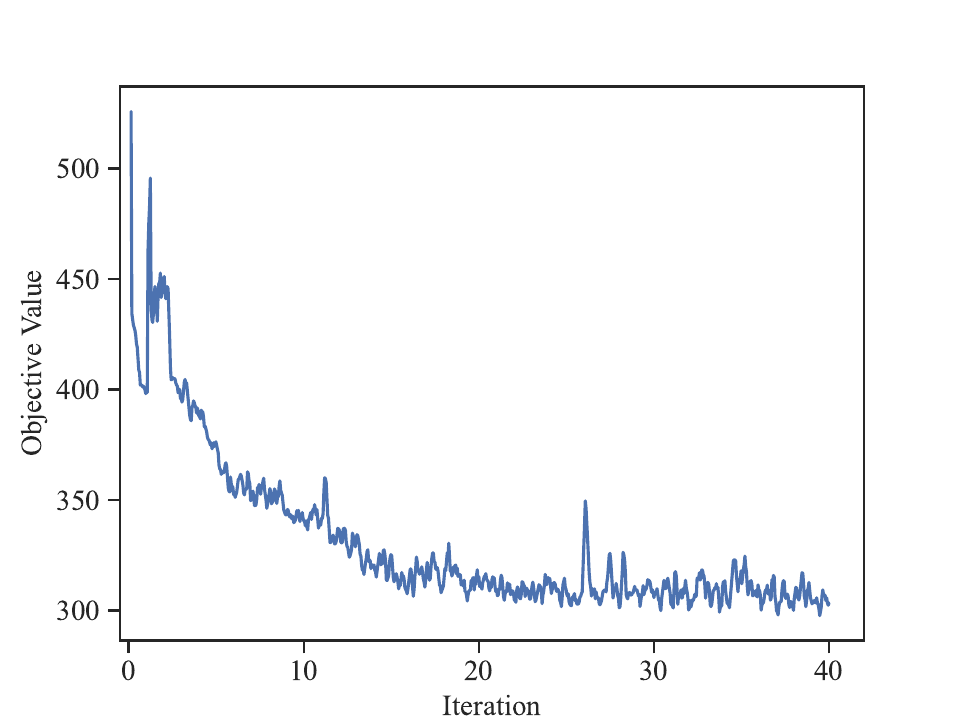}
        }
        \subfigure[S5]{
        \includegraphics[width=0.31\linewidth]{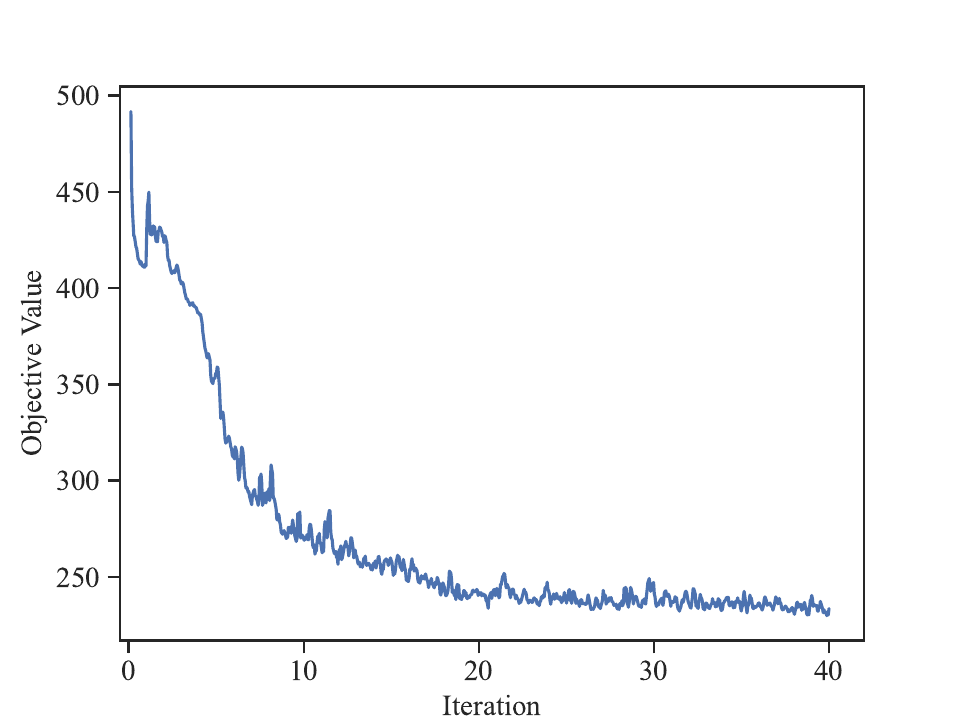}
        }
        \subfigure[S6]{
        \includegraphics[width=0.31\linewidth]{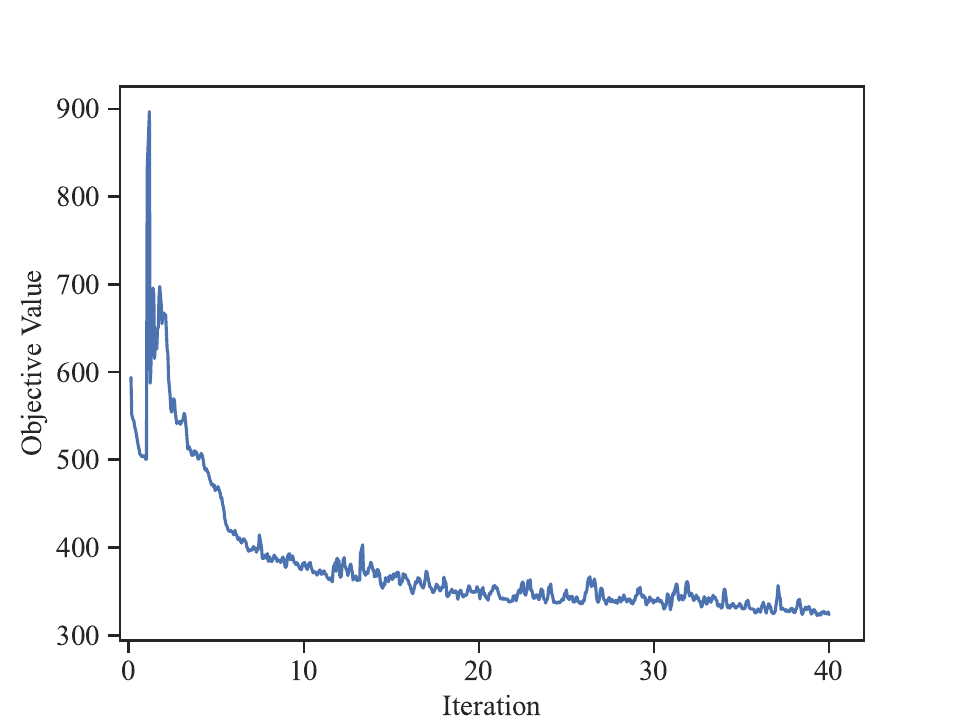}
        }
	\caption{Training Process of ADRL Models}
	\label{fig:5}
\end{figure}

Based on the experimental settings, we train ADRL models for each system, using the corresponding pre-trained MAM models as routing solvers. The learning curves for all the training models are shown in Figure \ref{fig:5}. On average, each training iteration consumes 222.81 min, 186.90 min, 222.67 min, 268.97 min, 250.23 min, and 257.12 min for systems from S1 to S6, respectively. While the learning curves exhibit fluctuations at the beginning of training, they ultimately converge stably, indicating that the ADRL models have successfully learned valid policies for TSA.

For SBP, we conducted a series of comprehensive experiments to identify reasonable parameter settings. We are interested in: 1) the time slot range of future customers, 2) the sampling horizon, and 3) the number of sampled scenarios $q$. We considered both finite and infinite time window ranges for future customers, sampling horizons of 1, 2, and 3 days, as well as 10, 30, and 50 scenario samples. The experiment results in Appendix C indicate that an infinite time slot range, a 1-day sampling horizon, and 30 scenario samples performed well across a variety of systems. Based on the results, we use these settings for the remaining experiments.

\subsection{Benchmark Policies}
We evaluate the performance of our approaches against four benchmark policies. Two of them follow a predefined assignment rule, utilizing the OR-Tools for routing, while the other two benchmarks are based on the rollout framework.

\subsubsection{Two Rule-based Policies}
In contrast to the proposed approaches, which utilize the rollout framework or scenario sampling for future approximation, the two rule-based policies directly assign time slots to customers based on the current state. These are referred to as myopic approaches and are commonly used as benchmarks in dynamic VRPs \citep{baty2024combinatorial}. The first policy, called the \textit{Random Policy} (RAN), randomly assigns service time slot to each arriving customer within the available time horizon. 

The second policy, called the \textit{Segmentation Policy} (SEG), utilizes a region segmentation idea for TSA. Specifically, the entire service region is divided into ten areas, each corresponding to a sector with equal angular width in radians, starting from the depot. These areas are labeled from 1 to 10 in a counterclockwise direction, beginning from the positive $x$-axis. The TSA decision for a customer depends on the region in which they are located. For example, on the first day, customers located in regions 1 and 2 are assigned to the morning and afternoon of the second day, respectively. However, one the second day, areas 1 and 2 correspond to TSA decisions for the seventh day, as all customers assigned to the second day are served on that day.

\subsubsection{Two Rollout-based Policies}
We integrate the RAN and SEG policies into the rollout framework to construct two rollout-based policies, termed RAN with rollout execution (RAN-RE) and SEG with rollout execution (SEG-RE), respectively. For RAN-RE, all the available time slots are considered as promising actions for the TSA decision. In addition, we replace the ADRL with RAN to calculate the objective values for the rollout dataset of these promising actions to determine the final TSA decision. Similarly, SEG-RE also considers all available time slots, but uses SEG to obtain objective values of promising actions.

% \subsection{Computational Results}
% In the following section, we evaluate the performance of our SBP and ADRL-RE by comparing it with benchmark policies. Then, we conduct detailed experiments for our ADRL-RE to validate the effectiveness of the rollout framework within ADRL-RE. Finally, we present an experiment conducted on a simulation derived from real-world data to assess the practical applicability of our approaches.

\subsection{Performance of the Methods} \label{sec: Comparison Analysis}
We begin by comparing the computational performance of our ADRL-RE and SBP against benchmark policies in terms of total cost, assignment efficiency, and service satisfaction. All approaches are tested in each system with 100 randomly generated instances, and the average value across these instances is reported. Bold-faced numbers in tables indicate the best-performing statistic.

% Table \ref{tab:4} shows the percentage of customers in which TSA decision accurately assigns a customer's preferred time window. Besides, we summarize the detailed statistics in Table \ref{tab:5} and \ref{tab:6}, where Table \ref{tab:5} lists the total travel cost (TTC) and delay penalty (DP) during the whole time horizon and Table \ref{tab:6} presents the standard deviation of the number of customers served each day to reflect the service equilibrium (SE).

% Our analysis focuses on the following evaluation metrics: 1) objective value (OV), the total cost of the solution; 2) assignment efficiency (AE), TSA decision time per epoch in seconds; 3) service equilibrium (SE), the standard deviation of the number of customers served each day; 4) travel cost (TC), including the traveling and waiting time of customers; 5) delay penalty (DP); and 6) customer satisfaction ratio (CSR), the percentage of customers in which TSA decision accurately assigns a customer's preferred time window. 

\begin{table}[htbp] \footnotesize
    \centering
    \caption{Total Cost of Each Approach}
    \renewcommand\arraystretch{1.2}
    \begin{tabular}{ccccccc}
        \toprule
        System & RAN & SEG & RAN-RE & SEG-RE & SBP & ADRL-RE \\ \midrule
        S1 & 415.31 & 327.17 & 284.63 & 194.05 & 234.76 & \textbf{176.88} \\
        S2 & 471.78 & 380.80 & 354.68 & 298.71 & 308.89 & \textbf{254.32} \\
        S3 & 545.17 & 457.00 & 444.33 & 404.19 & 375.29 & \textbf{335.41} \\
        S4 & 440.08 & 369.18 & 339.87 & 173.15 & 261.29 & \textbf{165.18} \\
        S5 & 493.43 & 411.00 & 362.11 & 282.58 & 319.72 & \textbf{243.63} \\
        S6 & 543.05 & 466.02 & 441.78 & 366.05 & 384.93 & \textbf{310.68} \\ \bottomrule
    \end{tabular} \label{tab:2}
\end{table}

\begin{table}[htbp] \footnotesize
    \centering
    \caption{TSA Decision Time per Epoch of Online Excution Approaches (s)}
    \renewcommand\arraystretch{1.2}
    \begin{tabular}{ccccc}
        \toprule
        System & RAN-RE & SEG-RE & SBP & ADRL-RE \\ \midrule
        S1 & \textbf{0.70} & 0.74 & 2.13 & 0.74 \\
        S2 & \textbf{0.70} & 0.75 & 2.03 & 0.71 \\
        S3 & \textbf{0.71} & 0.74 & 2.01 & 0.76 \\
        S4 & \textbf{0.87} & 0.91 & 2.08 & 0.97 \\
        S5 & \textbf{0.84} & 0.90 & 2.08 & 0.89 \\
        S6 & \textbf{0.87} & 0.90 & 2.08 & 0.93 \\ \bottomrule
    \end{tabular} \label{tab:3}
\end{table}

\begin{figure}[htbp]
	\centering
	\includegraphics[width=0.7\linewidth]{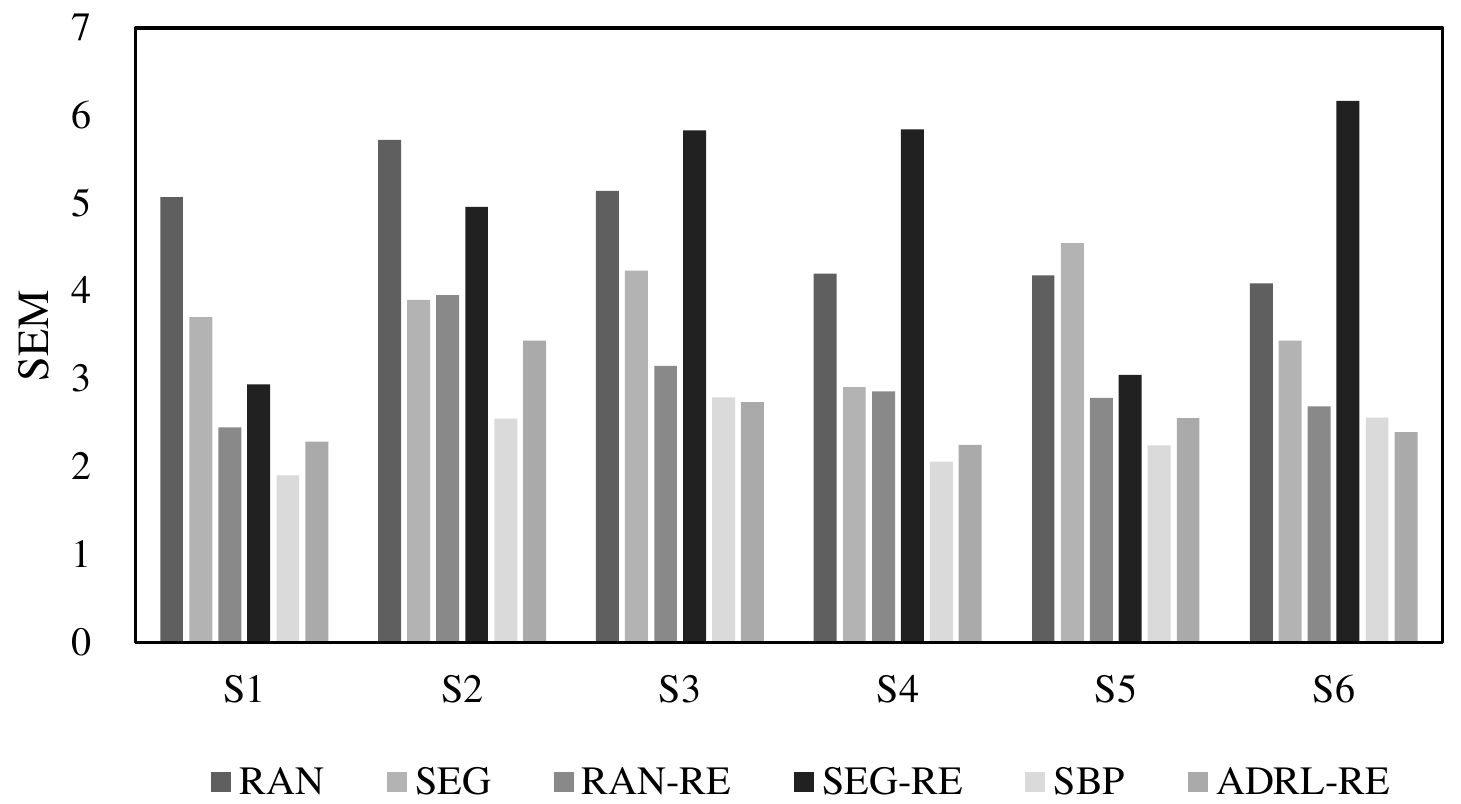}
	\caption{Standard Error of the Mean (SEM) of the Total Cost}
	\label{fig:6}
\end{figure}

Table \ref{tab:2} and Table \ref{tab:3} present each approach's total cost and TSA decision time per epoch, respectively. From Table \ref{tab:2}, rule-based policies RAN and SEG exhibit consistently high costs across all simulations. However, the integration of the rollout framework yields substantial performance enhancements, with RAN-RE and SEG-RE achieving average cost reductions of 23.80\% and 29.93\% compared to RAN and SEG, respectively. For the standard error of the mean (SEM) of total cost presented in Figure \ref{fig:6}, RAN-RE demonstrates a reduced SEM compared to the RAN, while SEG-RE exhibits a relatively higher SEM. One potential reason for this is that the RAN makes decisions randomly, resulting in significant instability. By anticipating future scenarios, RAN-RE reduces stochasticity by leveraging the solutions from rollout instances to guide decision-making. In contrast, SEG-RE maintains the inherent rigidity of its segmentation-based logic, where the rollout-enhanced version still follows to predefined region division, making it less robust and sensitive to scenario variations.

Regarding SBP, it exhibits superior performance compared to RAN, SEG, and RAN-RE. Although SBP is slightly trailing behind SEG-RE, it has a significantly lower SEM in total cost, indicating great reliability in various scenarios. However, SBP is the most computationally expensive, with an average of 2.07 seconds for each TSA decision epoch. Notably, ADRL-RE achieves the lowest total cost among all systems with relatively little decision time. Despite SEG-RE already offering high solution quality, ADRL-RE achieves an average improvement of 12.37\%. This enhancement is largely attributable to the effective TSA policy based on the rollout framework and well-trained ADRL models.

\begin{table}[htbp] \footnotesize
    \centering
    \caption{Satisfied Assignment Ratio (\%) of Each Approach}
    \renewcommand\arraystretch{1.2}
    \begin{tabular}{ccccccc}
        \toprule
        System & RAN & SEG & RAN-RE & SEG-RE & SBP & ADRL-RE \\ \midrule
        S1 & 30.31 & 29.86 & 51.78 & 78.48 & 57.66 & \textbf{90.53} \\
        S2 & 29.87 & 29.86 & 52.76 & 66.74 & 58.68 & \textbf{89.56} \\
        S3 & 30.18 & 29.86 & 46.42 & 59.97 & 58.28 & \textbf{83.27} \\
        S4 & 29.63 & 30.45 & 46.43 & 88.73 & 57.69 & \textbf{89.86} \\
        S5 & 29.63 & 30.45 & 53.67 & 73.95 & 58.07 & \textbf{85.17} \\
        S6 & 30.13 & 30.45 & 47.82 & 69.38 & 58.09 & \textbf{81.38} \\ \bottomrule
    \end{tabular} \label{tab:4}
\end{table}

\begin{table}[htbp] \footnotesize
    \centering
    \caption{Total Travel Cost (TTC) and Delay Penalty (DP) of Each Approach}
    \renewcommand\arraystretch{1.2}
    \begin{tabular}{ccccccrccrrcc}
        \toprule
        \multirow{2}[2]{*}{Simulation} & \multicolumn{2}{c}{RAN} & \multicolumn{2}{c}{SEG} & \multicolumn{2}{c}{RAN-RE} & \multicolumn{2}{c}{SEG-RE} & \multicolumn{2}{c}{SBP} & \multicolumn{2}{c}{ADRL-RE} \\
        \cmidrule(lr){2-3} \cmidrule(lr){4-5} \cmidrule(lr){6-7} \cmidrule(lr){8-9} \cmidrule(lr){10-11} \cmidrule(lr){12-13}
        & TTC & DP & TTC & DP & TTC & \multicolumn{1}{c}{DP} & TTC & DP & \multicolumn{1}{c}{TTC} & \multicolumn{1}{c}{DP} & TTC & DP \\
        \midrule
        S1 & 128.53 & 75.77 & \textbf{88.69} & 26.10 & 125.57 & 12.81 & 113.93 & 14.80 & 98.65 & \textbf{7.92} & 123.64 & 24.60 \\
        S2 & 189.53 & 69.87 & \textbf{140.65} & 27.78 & 180.34 & 31.12 & 177.53 & 20.36 & 162.42 & \textbf{21.49} & 175.08 & 47.62 \\
        S3 & 266.68 & 67.15 & \textbf{211.27} & 33.35 & 266.74 & \textbf{15.21} & 254.91 & 27.92 & 231.84 & 16.97 & 257.42 & 27.36 \\
        S4 & 117.53 & 41.98 & \textbf{78.06} & 13.82 & 113.30 & 12.51 & 101.50 & 26.51 & 86.85 & \textbf{6.91} & 110.73 & 13.99 \\
        S5 & 172.05 & 43.78 & \textbf{119.06} & 14.64 & 163.94 & 13.51 & 157.59 & 20.95 & 142.29 & \textbf{10.32} & 159.90 & 24.60 \\
        S6 & 229.21 & 35.21 & \textbf{172.03} & 16.69 & 225.42 & \textbf{8.22} & 217.19 & 26.76 & 202.69 & 15.26 & 220.28 & 16.22 \\
        \bottomrule
    \end{tabular} \label{tab:5}
\end{table}

Next, we look into the detailed statistics for further analysis. Table \ref{tab:4} reports the satisfied assignment ratio (SAR) achieved by each approach, defined as the percentage of customers whose TSA decisions fall within their preferred time slots. ADRL-RE always achieves the highest SAR, with a 20.36\% improvement compared to SEG-RE. Although SBP has lower satisfied assignment ratio than SEG-RE, it remains largely unaffected by problem size, number of vehicles, and travel coefficient.

Table \ref{tab:5} reports the total traveling cost and delay penalty to reflect the quality of the routing solutions. On the one hand, we see that SBP generally achieves a lower delay penalty, indicating its ability to meet the assigned time slots as closely as possible during actual service to customers. On the other hand, ADRL-RE shows higher overall travel cost and delay penalty than SEG-RE and SBP. This feels intuitive, as a higher satisfied assignment ratio leads to more customers being assigned their preferred time slots, thus reducing the temporal flexibility in route optimization. However, when considering the standard deviation of the number of customers served each day in Table \ref{tab:6}, we see that ADRL-RE has a notably low standard deviation, indicating it effectively balances the number of customers served daily.

\begin{table}[htbp] \footnotesize
    \centering
    \caption{Standard Deviation of Number of Customer Served Each Day}
    \renewcommand\arraystretch{1.2}
    \begin{tabular}{ccccccc}
        \toprule
        Simulation & RAN & SEG & RAN-RE & SEG-RE & SBP & ADRL-RE \\
        \midrule
        S1 & 5.20 & 5.32 & 3.30 & 3.95 & 4.80 & \textbf{3.66} \\
        S2 & 5.21 & 5.32 & 5.17 & \textbf{3.92} & 5.22 & 4.90 \\
        S3 & 5.20 & 5.32 & 3.39 & 3.86 & 5.32 & \textbf{3.29} \\
        S4 & 6.34 & 6.57 & 4.91 & 5.44 & 6.67 & \textbf{4.80} \\
        S5 & 6.47 & 6.57 & 6.04 & \textbf{5.41} & 6.88 & 6.05 \\
        S6 & 6.40 & 6.57 & 4.72 & 5.17 & 7.27 & \textbf{4.25} \\
        \bottomrule
    \end{tabular} \label{tab:6}
\end{table}
       
% \begin{figure}[htbp]
% 	\centering
%         \subfigure[S1]{
%         \includegraphics[width=0.31\linewidth]{figure/box/S1.pdf}
%         }
%         \subfigure[S2]{
%         \includegraphics[width=0.31\linewidth]{figure/box/S2.pdf}
%         }
%         \subfigure[S3]{
%         \includegraphics[width=0.31\linewidth]{figure/box/S3.pdf}
%         }
%         \subfigure[S4]{
%         \includegraphics[width=0.31\linewidth]{figure/box/S4.pdf}
%         }
%         \subfigure[S5]{
%         \includegraphics[width=0.31\linewidth]{figure/box/S5.pdf}
%         }
%         \subfigure[S6]{
%         \includegraphics[width=0.31\linewidth]{figure/box/S6.pdf}
%         }
% 	\caption{Comparison of Objective Value for Four Approaches Across 6 Simulations and 100 Instances Each}
% 	\label{fig:9}
% \end{figure}

% For a more intuitive comparison, Figure \ref{fig:9} shows that R-ADRL* exhibits a lower median and interquartile range compared to SBP, demonstrating its superior and more stable performance across all simulations. Despite this, SBP tends to have fewer outliers, particularly in simulations S4 and S6. This phenomenon relates to the scenario-sampling approach on which SBP is based, providing a robust solution that performs across all instances. Our R-ADRL* approach, on the other hand, balances the trade-off between robustness and solution quality. Additionally, the utilization of batch processing and the efficient inference of DRL substantially leads to a significant reduction in computation time compared to SBP.

In summary, ADRL-RE outperforms other benchmarks in terms of both total cost and service satisfaction, although it comes with higher travel cost and delay penalty. ADRL-RE gives out robust and high-quality solutions over a variety of systems. In contrast, RAN and SEG show the worst performance, indicating that rigid rule-based logic likely fails to adapt to complex and dynamic scenarios. However, the rollout framework can be effective in compensating for the deficiencies of static rule policies, highlighting the significance of incorporating future information when making sequential decisions in a dynamic nature. Meanwhile, SBP offers moderate performance, providing a balanced and stable solution regarding travel cost, delay penalty, and service satisfaction.

\subsection{Extended analysis of ADRL-RE}
The comparison of benchmarks reveals that ADRL-RE performs best in terms of total cost. In this section, further experiments are conducted to evaluate the configuration of ADRL-RE, and a detailed analysis is provided regarding the effectiveness of the rollout framework and promising action selection strategy.

\subsubsection{Rollout Framework} \label{sect: Rollout Framework}
Unlike end-to-end DRL approaches in combinatorial optimization that directly generate the solution \citep{wang2024end, li2024multi}, our ADRL-RE approach integrates the rollout framework to enhance solution quality. To verify its effectiveness, we compare our ADRL-RE with the end-to-end ADRL (EE-ADRL), and the results are illustrated in Table \ref{tab:7}.

\begin{table}[htbp] \footnotesize
    \centering
    \caption{Results for EE-ADRL and ADRL-RE}
    \renewcommand\arraystretch{1.2}
    \begin{tabular}{ccccccccc}
        \toprule
        \multirow{2}[2]{*}{Simulation} & \multicolumn{4}{c}{EE-ADRL} & \multicolumn{4}{c}{ADRL-RE} \\
        \cmidrule(lr){2-5} \cmidrule(lr){6-9}
        & TC & SAR & SE & DT & TC & SAR & SE & DT \\
        \midrule
        S1 & 272.372 & 58.83 & 3.63 & 0.00 & 176.88 & 90.53 & 3.66 & 0.74 \\
        S2 & 282.006 & 58.68 & 4.58 & 0.00 & 254.32 & 89.56 & 4.90 & 0.71 \\
        S3 & 459.871 & 44.24 & 3.23 & 0.00 & 335.41 & 83.27 & 3.29 & 0.76 \\
        S4 & 346.212 & 42.77 & 5.09 & 0.00 & 165.18 & 89.86 & 4.80 & 0.97 \\
        S5 & 393.133 & 49.76 & 6.18 & 0.00 & 243.63 & 85.17 & 6.05 & 0.89 \\
        S6 & 403.373 & 56.99 & 2.79 & 0.00 & 310.68 & 81.38 & 4.25 & 0.93 \\
        \bottomrule
    \end{tabular} \label{tab:7}
    \vspace{0.3cm}
    \par \raggedright
    \textit{Note}: TC denotes the total cost, SAR the satisfied assignment ratio, SE the standard deviation of the number of customers served each day, and DT the TSA decision time per epoch.
\end{table}

From Table \ref{tab:7}, it is evident that the rollout framework significantly enhances solution quality, with the most notable improvement of 52.29\% in system S4. To further investigate this, we analyze the selection probabilities at each TSA decision epoch. In some cases, the probabilities for multiple actions are very close. With the use of the greedy strategy, EE-ADRL always selects the action with the highest probability based on the currently available information. However, an action with a slightly lower probability may, in fact, represent a better decision when considering future information. The rollout framework generates several rollout instances by approximating potential future scenarios and then solves these to determine the final decision. In this way, we leverage future information to identify actions that may not seem optimal initially but lead to better outcomes when the long-term impact is considered. 

Based on the above analysis, we note that the ADRL models trained for each system have not fully converged, as indicated by the downward trend in their training curves shown in Figure \ref{fig:5}. This incomplete convergence results in the learned policies to be "wavering", which can lead to poor decision-making. Although additional training could mitigate this issue, the huge time cost associated with further training becomes a limiting factor, especially considering the already slow convergence rate towards the end of the training curves. In this context, the rollout framework plays a vital role in improving decision quality without requiring extensive training time.

\subsubsection{Different Strategies of Promising Action Selection}
We now investigate the performance of different promising action selection strategies within the rollout framework. Specifically, we compare with ADRL-RE-A, which considers all the available time slots as promising actions.

\begin{table}[htbp] \footnotesize
    \centering
    \caption{Results for ADRL-RE-A and ADRL-RE}
    \renewcommand\arraystretch{1.2}
    \begin{tabular}{ccccccccc}
        \toprule
        \multirow{2}[2]{*}{Simulation} & \multicolumn{4}{c}{ADRL-RE-A} & \multicolumn{4}{c}{ADRL-RE} \\
        \cmidrule(lr){2-5} \cmidrule(lr){6-9}
        & TC & SAR & SE & DT & TC & SAR & SE & DT \\
        \midrule
        S1 & 200.65 & 92.86 & 4.92 & 0.78 & 176.88 & 90.53 & 3.66 & 0.74 \\
        S2 & 250.10 & 89.63 & 4.95 & 0.74 & 254.32 & 89.56 & 4.90 & 0.71 \\
        S3 & 352.64 & 81.35 & 4.20 & 0.77 & 335.41 & 83.27 & 3.29 & 0.76 \\
        S4 & 165.72 & 90.71 & 5.29 & 0.98 & 165.18 & 89.86 & 4.80 & 0.97 \\
        S5 & 222.84 & 88.46 & 4.74 & 0.92 & 243.63 & 85.17 & 6.05 & 0.89 \\
        S6 & 319.86 & 80.19 & 5.13 & 0.96 & 310.68 & 81.38 & 4.25 & 0.93 \\
        \bottomrule
    \end{tabular} \label{tab:8}
    \vspace{0.3cm}
    \par \raggedright
    \textit{Note}: TC denotes the total cost, SAR the satisfied assignment ratio, SE the standard deviation of the number of customers served each day, and DT the TSA decision time per epoch.
\end{table}

Table \ref{tab:8} presents the overall performance of ADRL-RE-A and ADRL-RE. We see that ADRL-RE generally has a lower total cost, with an average reduction of 1.49\% compared to ADRL-RE-A. Although ADRL-RE-A achieves slightly higher service satisfaction, particularly in S1 and S4, it also incurs a higher standard deviation in the number of customers served per day, reflecting poorer service equilibrium. Besides, ADRL-RE-A exhibits longer decision times due to considering all available actions, whereas ADRL-RE selectively considers only some of these actions. The results suggest that ADRL-RE, considering promising actions with non-zero probability, has a better overall performance, which strikes a better balance among cost, customer satisfaction, service equilibrium, and time efficiency.

\subsection{Case Study}
\begin{figure}[htbp]
	\centering
	\includegraphics[width=0.55\linewidth]{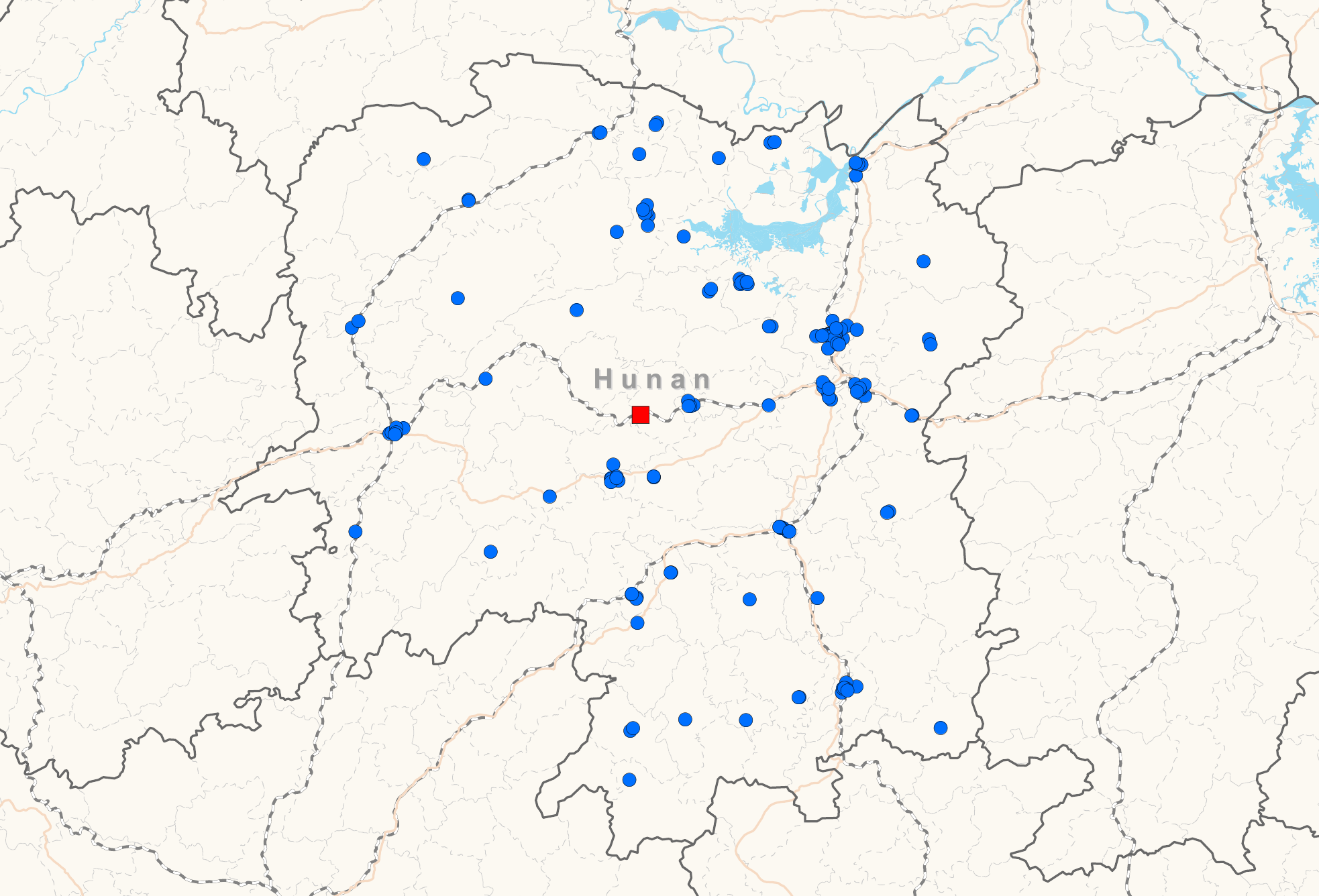}
	\caption{Grade II and above medical institutions in Hunan Province, with the depot in red and the institutions in blue.}
	\label{fig:7}
\end{figure}

To further evaluate the applicability of proposed approaches, we generated a system related to the after-sales service of large medical equipment in Hunan Province, China. Hunan covers an area of 211,800 square kilometers and has 161 medical institutions that are classified as grade II or above, as presented in Figure \ref{fig:7}. In this system, a centrally located large medical equipment supplier serves as the depot, and the system parameters align with those in S2 as outlined in Table \ref{tab:1}. The customer coordinates are sampled from the locations of medical institutions, and the average performance of the approaches over 20 runs is recorded in Table \ref{tab:9}.

Three observations stand out. First, ADRL-RE still outperforms benchmark policies, with the lowest total cost and highest satisfied assignement ratio. This reflects its excellent after-sales service from both the TSA and RP processes. Second, unlike the results in Section \ref{sec: Comparison Analysis}, where SEG has better performance than RAN, SEG is much worse than RAN here. This highlights the challenge of relying solely on geographical location for TSA, as it struggles to adapt to the variability in customer distribution. However, the rollout framework can efficiently alleviate this issue as SEG-RE has good performance. Third, SBP still has moderate performance with a similar SAR in Section \ref{sec: Comparison Analysis}, indicating its stability in a variety of scenarios. Overall, our ADRL-RE shows strong adaptability to different customer distributions and ensures its practicality across diverse scenarios.

\begin{table}[htbp] \footnotesize
  \centering
  \caption{Results of the Case Study}
  \renewcommand\arraystretch{1.2}
  \begin{tabularx}{0.5\textwidth}{l>{\centering\arraybackslash}X >{\centering\arraybackslash}X >{\centering\arraybackslash}X >{\centering\arraybackslash}X}
    \toprule
        {Approach} & TC & SAR & SE & DT \\
        \midrule
        RAN & 355.02 & 28.67 & 4.71 & 0.00 \\
        SEG & 453.02 & 32.17 & 8.75 & 0.00 \\
        RAN-RE & 290.43 & 45.35 & 4.38 & 0.70 \\
        SEG-RE & 185.76 & 79.83 & 3.82 & 0.67 \\
        SBP & 236.99 & 59.48 & 5.72 & 2.03 \\
        ADRL-RE & 143.66 & 93.92 & 4.79 & 0.69 \\
        \bottomrule
  \end{tabularx} \label{tab:9}
    \vspace{0.3cm}
    \par \raggedright
    \textit{Note}: TC denotes the total cost, SAR the satisfied assignment ratio, SE the standard deviation of the number of customers served each day, and DT the TSA decision time per epoch.
\end{table}%

% Overall, by incorporating a wide range of features such as customer location, appearance time, preferred and assigned time windows, our ADRL-RE shows strong adaptability to different customer distributions and ensuring its practicality across diverse scenarios.

\section{Conclusion} \label{sec: CON}
In this paper, we introduce the dynamic time slot assignment problem with commitments and customer preferences (DTSAP-CCP) in the context of after-sales service. In particular, we consider an OEM active in the business-to-business context that needs to make time slot assignment decisions in real-time during the communication with customers, thus having only little time for computations. Notably, customers have their preferred time slots that the OEM has to consider while also minimizing total costs. At the end of each day, serving routes for customers assigned to the next day should be determined.

We model the problem as a finite-horizon MDP and propose two distinct approaches: 1) an attention-based deep reinforcement learning approach with a rollout execution (ADRL-RE), and 2) a scenario-based planning approach (SBP) based on a well-established scenario sampling framework. For ADRL-RE, an attention-based DRL model is constructed and offline trained first. A rollout framework with online trajectory simulations for promising actions is utilized in the actual determination of TSA decisions. Regarding SBP, it samples several scenarios upon the arrival of each customer, where a time slot assignment decision is guided by the solution of these sampled scenarios. 

Our computational experiments on a variety of systems show that ADRL-RE is effective in obtaining high-quality solutions, with significant advantages in comparison with both two rule-based and two rollout-based approaches. SBP only achieves moderate performance that is inferior to SEG-RE and ADRL-RE, while it shows great stability with less SEM of total cost in most simulations. Other extended experiments are conducted for the configuration of ADRL-RE. The results suggest that considering actions with non-zero selection probability instead of all available can achieve a better overall performance, especially with a large vehicle fleet. Additionally, the adaptability for out-of-distribution instances is also verified by case study.

Whereas ADRL-RE yields well outcomes, it also presents limitations that call for further exploration and improvement. For instance, the training time for ADRL models is relatively high. A more rigorous hyperparameter tuning and a more effective reinforcement learning training algorithm design could help mitigate this issue. Additionally, the current approach assumes relatively stable environments, whereas real-world applications often involve unpredictable changes in customer behavior or external factors. Integrating adaptive learning strategies or incorporating real-time data updates could improve the model's responsiveness and robustness. Another intriguing extension would involve improving adaptability to different temporal and spatial distributions of customers, which would enable the model to better handle dynamic environments with varying demand patterns. Techniques like transfer learning could be helpful, allowing the trained model to generalize more effectively across different systems.

\bibliographystyle{informs2014trsc} % outcomment this and next line in Case 1
\bibliography{Main} % if more than one, comma separated

%\THEEndNotes
% \begingroup \parindent 0pt \parskip 0.0ex \def\enotesize{\normalsize} \theendnotes \endgroup

% Appendix here
% Options are (1) APPENDIX (with or without general title) or
%             (2) APPENDICES (if it has more than one unrelated sections)
% Outcomment the appropriate case if necessary
%
% \begin{APPENDIX}{<Title of the Appendix>}
% \end{APPENDIX}
%
%   or
%
\begin{APPENDICES}
\section{Modules of ADRL}

\subsection{Pre-Norm Attention Layer}
The Pre-Norm attention layer is consists of a multi-head attention (MHA) sub-layer and a feed-forward network (FFN) sub-layer. Different from the Post-Norm attention layer \citep{vaswani2017attention}, which follows the order "MHA/FFN sub-layer $\rightarrow$ skip connection $\rightarrow$ normalization", the Pre-Norm attention layer applies normalization ahead of the sub-layers as shown in Figure \ref{fig:8}, improving training stability \citep{xiong2020layer, hesimplifying}. Let $H_k^{l-1}$ denote the input of the $l$-th attention layer, which is first processed by the normalization. Following \citet{koolattention}, we adopt an 8-heads self-attention mechanism in the MHA sub-layer, and the updated node embeddings $\hat{H}_k^{l}$ are obtained via a skip connection. The above processes are formally defined as follows:
\begin{equation} \label{eq_qkv}
    q_i = \Tilde{H}_k^{l}W^q_i, k_i = \Tilde{H}_k^{l}W^k_i, v_i = \Tilde{H}_k^{l}W^v_i,
\end{equation}
\begin{equation}
    \mathrm{head_i} = \mathrm{softmax}(\frac{q_i^T \cdot k_i}{\sqrt{dim}}) \cdot v_i,
\end{equation}
\begin{equation}
    \mathrm{MHA}(H_k^{l-1}) = \mathrm{Concat}(\mathrm{head_1}, \cdots, \mathrm{head_8})W^c,
\end{equation}
\begin{equation}
    \hat{H}_k^{l} = H_k^{l-1} + \mathrm{MHA}(H_k^{l-1}),
\end{equation}
where $\Tilde{H}_k^{l}$ is the normalized $H_k^{l-1}$, $dim$ is the embedding dimension, and $W^q_i$, $W^k_i$, $W^v_i$ and $W^c$ are learnable parameters.

\begin{figure}[htbp]
    \begin{minipage}[t]{0.5\linewidth}
        \centering
        \includegraphics[width=0.45\textwidth]{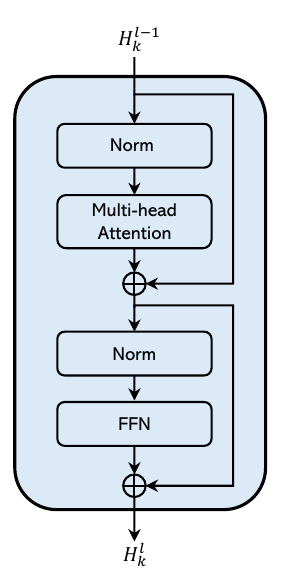}
        \caption{Pre-Norm Attention Layer}
        \label{fig:8}
    \end{minipage}%
    \begin{minipage}[t]{0.5\linewidth}
        \centering
        \includegraphics[width=0.7\textwidth]{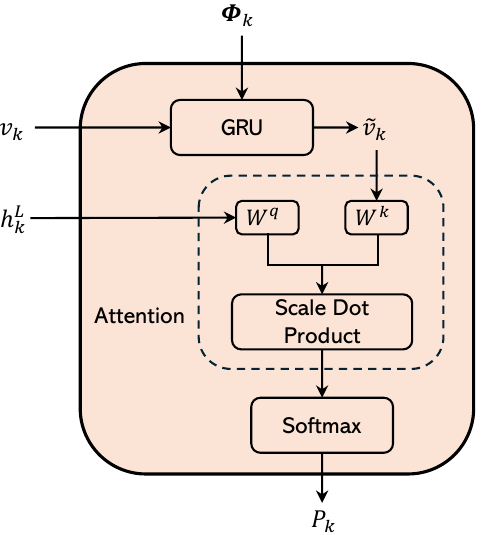}
        \caption{GRU-based Attention Layer}
        \label{fig:9}
    \end{minipage}
\end{figure}

Subsequently, the node embeddings $\hat{H}_k^{l}$ are processed further by the normalization operation and FFN sub-layer. The FFN sub-layer is constructed by two linear projections with a dimension of 512 and a ReLU activation function. Finally, the output of the $l$-th attention layer $H_k^{l}$ is derived through a skip connection as follows:
\begin{equation}
    H_k^{l} = \hat{H}_k^{l} + \mathrm{FFN}(\mathrm{Norm}(\hat{H}_k^{l})).
\end{equation}

\subsection{GRU-based Attention Layer}
This network module learns to obtain the selection probabilities for available assignment decisions by leveraging a GRU \citep{parisotto2020stabilizing} and a single-head attention operator, as shown in Figure \ref{fig:9}. The GRU receives the initial hidden vector $v_k$ and TSA embedding $\Phi_k$ as inputs. These inputs are used to generate the next hidden vector $\Tilde{v}_k$, which captures the temporal dependencies. For the attention operator, the query and key are derived through the multiplication of trainable parameters $W^q$ and $W^k$ with the assigning node embedding $h_k^L$ and hidden vector $\Tilde{v}_k$, respectively. The computation of the final selection probabilities proceeds as follows:
\begin{equation}
    \Tilde{v}_k = \mathrm{GRU}(v_k, C_k),
\end{equation}
\begin{equation}
    P_k = \mathrm{softmax}(\frac{(W^q h_k^L)^T \cdot W^k \Tilde{v}_k}{\sqrt{dim}}).
\end{equation}

At decision epoch $k$, a new customer arrives, and the current state is encoded and decoded once to obtain the selection probabilities for the available decisions. In this case, the next hidden vector $\Tilde{v}_k$ is used only for the attention operator. Since the encoding process is relatively time-consuming, during model training, we encode the new customers appearing on the same day once and then iteratively decode for them. Thus, $\Tilde{v}_k$ will also be utilized and updated through the GRU for the next decision epoch.

\section{Decoders of MAM}
The original AM demonstrates strong potential in solving vehicle routing problems, while the time windows of customers are not considered \citep{koolattention}. To efficiently solve the VRPSTW, we modify the AM by incorporating the Pre-Norm attention layer into the encoder and constructing two decoders: one for vehicle selection and the other for customer selection. The encoder of the MAM takes the node information of the depot and customers as inputs, with each node characterized by the coordinates and time window. The network architecture of the encoder resembles that of the original AM, with the only modification being the replacement of the Post-Norm attention layer with the Pre-Norm attention layer. After being processed by the encoder, the final node embeddings $U$ are obtained, along with the corresponding graph embedding $g_U$, which is derived by applying an average pooling operation on $U$. For decoding, two decoders are constructed: 1) vehicle selection decoder, which determines the vehicle that will visit the next customer, and 2) node selection decoder, which selects the next customer to be visited. These two decoders generate the vehicle-customer visiting sequences step by step, and the details of them are provided in the following sections.
\subsection{Vehicle Selection Decoder}
The vehicle selection decoder receives the route and vehicle information to output a probability distribution for selecting a specific vehicle. Two feature embeddings are constructed and leveraged: 1) route feature embedding, which extracts information from existing partial routes of all vehicles, and 2) vehicle feature embedding, which captures the state of vehicles at the current decoding step.
\begin{itemize}
    \item \textit{Route Feature Embedding $R_k$:} At the decoding step $k$, we define a route feature context $R_k^i$ for the vehicle $v_k^i$ as an arrangement of its visited node embeddings as follows:
    \begin{equation}
        R_k^i = \mathrm{Concat}(\hat{u}_0^i, \cdots, \hat{u}_{t-1}^i),
    \end{equation}
    where $\hat{u}_0^i$ is the node embedding of the first visited node for vehicle $v_k^i$. To obtain the route feature embedding $R_k$, we aggregate the route feature contexts of all vehicles by leveraging a max pooling operation and then concatenate them as follows:
    \begin{equation}
        R_k = \mathrm{Concat}(\mathrm{max}(R_k^1), \cdots, \mathrm{max}(R_k^p)),
    \end{equation}
    where $p$ is the vehicle quantity.

    \item \textit{Vehicle Feature Embedding $V_k$:} The vehicle feature embedding is derived from the vehicle feature context that consists of the vehicle locations and departure times. At the decoding step $k$, the feature of the vehicle $v_k^i$ includes its coordinates and the service end time of the last node. We construct a vehicle feature context $\hat{V}_k$ by concatenating the feature of all vehicles as follows:
    \begin{equation}
        \hat{V}_k = \mathrm{Concat}([vc_{k-1}^1, vd_{k-1}^1], \cdots, [vc_{k-1}^p, vd_{k-1}^p]),
    \end{equation}
    where $vc_{k-1}^1$ is the 2-dimensional location of the last node and $vd_{k-1}^1$ is the service end time of the last node of the vehicle $v_k^1$. After that, the vehicle feature context is linearly projected with learning parameters $W_1$ and $b_1$, and then processed by a FFN to engender the vehicle feature embedding at step $k$ as follows:
    \begin{equation}
        V_k = \mathrm{FFN}(W_1\hat{V}_k + b_1).
    \end{equation}
\end{itemize}

Based on the route feature embedding $R_k$ and vehicle feature embedding $V_k$, we cooperate the graph embedding $g_U$ to obtain the probability distribution via a single-head attention layer.The key vector for the attention mechanism $k_k^v$, is obtained by concatenating $R_k$ and $V_k$, followed by a linear projection, and we deem the $g_U$ as the query vector. The output of the single-head attention layer $O_k^v$ is computed as follows:
    \begin{equation}
        k_k^v = W_2 \mathrm{Concat}(V_k, R_k) + b_2,
    \end{equation}
    \begin{equation}
        O_k^v = C \cdot \mathrm{tanh}(\frac{g_U^T k_k^v}{\sqrt{\mathrm{dim}}}),
    \end{equation}
where $W_2$ and $b_2$ are learning parameters, and $C$ is set to 10 to scale the results. Finally, the probability distribution for vehicle selection at decoding step $t$ is engendered by employing the softmax operation on $O_k^v$. The selected vehicle $\hat{v}_k$ is then determined using either a greedy or sampling strategy. 
%The greedy strategy always selects the vehicle with the maximum probability, whereas the sampling strategy selects the vehicle according to its probability.

\subsection{Node Selection Decoder}
The node selection decoder receives the node embeddings $U$ from the encoder and the selected vehicle $\hat{v}_k$ from the vehicle selection decoder as inputs to generate a probability distribution over all unvisited nodes. At the decoding step $k$, a context vector ${N}_k$ is firstly engendered by the graph embedding $g_U$, the last node embedding of the selected vehicle $\hat{u}_{k-1}$, and the service end time of the last node for the selected vehicle $\hat{vd}_{k-1}$ as follows:
\begin{equation}
    {N}_k = g_U + W_3 \mathrm{Concat}(\hat{u}_{k-1}, \hat{vd}_{k-1}) + b_3,
\end{equation}
where $W_3$ and $b_3$ are learnable parameters.

After that, an MHA layer is applied, where the context vector ${N}_k$ serves as the query vector and the key and value vectors come from the node embeddings $U$, as follows:
\begin{equation}
    \hat{N}_k = \mathrm{MHA}({N}_k, W_k U, W_v U),
\end{equation}
where $W_k$ and $W_v$ are learning parameters. Finally, we engender the compatibility by comparing the relationship between the enhanced context $\hat{N}_k$ and the node embeddings $U$ and then generate the probability distribution $P_k^N$ as follows:
\begin{equation}
    O_k^n = C \cdot \mathrm{tanh}(\frac{q_n^T k_n}{\sqrt{\mathrm{dim}}}),
\end{equation}
\begin{equation}
    P_k^N = \mathrm{softmax}(O_k^n),
\end{equation}
where $q_n$ and $k_n$ are derived from the linear projection of $\hat{N}_k$ and $U$, respectively. Similar to the vehicle selection decoder, greedy or sampling strategy is applied to determine the next visit node for the vehicle $\hat{v}_k$.

\begin{table}[htbp] \footnotesize
  \centering
  \caption{Results of Different Sampling Horizons and Time Window Range in SBP}
  \renewcommand\arraystretch{1.2}
  \begin{tabular}{ccccccc}
        \toprule
        Simulation & Sampling Horizon & Approach & TC    & SAR   & SE   & DT   \\ \midrule
        S1         & 1                & SBP-inf  & \textbf{234.76} & 57.66 & 4.80 & 2.13 \\
                   &                  & SBP-f    & 260.95 & 54.87 & 4.03 & 2.03 \\
                   & 2                & SBP-f    & 261.42 & 54.20 & 3.88 & 2.04 \\
                   & 3                & SBP-f    & 262.37 & 53.76 & 3.60 & 2.06 \\
        S2         & 1                & SBP-inf  & \textbf{308.89} & 58.68 & 5.22 & 2.03 \\
                   &                  & SBP-f    & 323.48 & 54.46 & 4.13 & 2.03 \\
                   & 2                & SBP-f    & 324.19 & 53.75 & 3.93 & 2.04 \\
                   & 3                & SBP-f    & 324.34 & 53.57 & 3.65 & 2.07 \\
        S3         & 1                & SBP-inf  & \textbf{375.29} & 58.25 & 5.32 & 2.01 \\
                   &                  & SBP-f    & 402.07 & 54.02 & 4.17 & 2.04 \\
                   & 2                & SBP-f    & 401.06 & 53.52 & 4.02 & 2.04 \\
                   & 3                & SBP-f    & 401.24 & 53.55 & 4.01 & 2.04 \\
        S4         & 1                & SBP-inf  & \textbf{261.29} & 57.96 & 6.67 & 2.01 \\
                   &                  & SBP-f    & 290.61 & 54.30 & 5.13 & 2.03 \\
                   & 2                & SBP-f    & 292.62 & 53.61 & 4.86 & 2.06 \\
                   & 3                & SBP-f    & 294.52 & 53.29 & 5.23 & 2.05 \\
        S5         & 1                & SBP-inf  & \textbf{319.72} & 58.07 & 6.88 & 2.08 \\
                   &                  & SBP-f    & 343.81 & 53.88 & 5.14 & 2.04 \\
                   & 2                & SBP-f    & 345.17 & 53.79 & 4.93 & 2.04 \\
                   & 3                & SBP-f    & 347.50 & 53.26 & 4.49 & 2.07 \\
        S6         & 1                & SBP-inf  & \textbf{384.93} & 58.14 & 7.27 & 2.08 \\
                   &                  & SBP-f    & 403.17 & 56.26 & 5.24 & 2.04 \\
                   & 2                & SBP-f    & 403.63 & 53.61 & 4.81 & 2.08 \\
                   & 3                & SBP-f    & 407.07 & 53.29 & 4.61 & 2.10 \\ \bottomrule
    \end{tabular}
    \vspace{0.3cm}
    \par \raggedright
    \textit{Note}: Bold refers to lowest total cost. TC denotes the total cost, SAR the satisfied assignment ratio, SE the standard deviation of the number of customers served each day, and DT the TSA decision time per epoch.
  \label{tab:10}%
\end{table}%

\section{Results of Parameter Setting Experiments for SBP}
To determine the appropriate sampling horizon and time slot range of future customers in SBP, a comparison experiment is conducted, and the results are reported in Table \ref{tab:10}. The approach implemented with an infinite time slot range is denoted as SBP-inf, while SBP-f refers to the approach with a finite time slot range for future customers. The results show that SBP-inf with a 1-day sampling horizon can always achieve the lowest total cost and the highest satisfied assignment ratio. 

\begin{table}[htbp] \footnotesize
  \centering
  \caption{Results of Different Sampling Size in SBP}
  \renewcommand\arraystretch{1.2}
  \begin{tabular}{ccccc}
        \toprule
        Sampling Size & TC    & SAR   & SE   & DT   \\ \midrule
        10            & 309.90 & 58.20 & 5.36 & 2.01 \\
        30            & 308.89 & 58.68 & 5.22 & 2.03 \\
        50            & 315.47 & 57.59 & 5.40 & 4.03 \\ \bottomrule
    \end{tabular}
    \vspace{0.3cm}
    \par \raggedright
    \textit{Note}: TC denotes the total cost, SAR the satisfied assignment ratio, SE the standard deviation of the number of customers served each day, and DT the TSA decision time per epoch.
  \label{tab:11}%
\end{table}%

Based on the above results, we further evaluate the SBP-inf with a 1-day sampling horizon under different sampling sizes in system S2. The results are listed in Table \ref{tab:11}, where the best performance is achieved with 30 scenario samples.

\section{Results of Different Rollout size in ADRL-RE}
\begin{table}[htbp] \footnotesize
  \centering
  \caption{Results of Different Rollout size in ADRL-RE}
  \renewcommand\arraystretch{1.2}
  \begin{tabular}{ccccc}
        \toprule
        Rollout Size & TC    & SAR   & SE   & DT   \\ \midrule
        5            & 192.41 & 79.32 & 3.20 & 0.71 \\
        10           & 176.88 & 90.53 & 3.66 & 0.74 \\
        20           & 163.65 & 87.92 & 3.14 & 0.82 \\ \bottomrule
    \end{tabular}
    \vspace{0.3cm}
    \par \raggedright
    \textit{Note}: TC denotes the total cost, SAR the satisfied assignment ratio, SE the standard deviation of the number of customers served each day, and DT the TSA decision time per epoch.
  \label{tab:12}%
\end{table}%

\end{APPENDICES}

% Acknowledgments here
% \ACKNOWLEDGMENT{We would like to express our sincere gratitude to [acknowledge individuals, organizations, or institutions] for their invaluable contributions to this research. We are also grateful to [mention any additional acknowledgements, such as technical assistance, data providers, or colleagues] for their support and assistance throughout the course of this work.}

% References here (outcomment the appropriate case)

% CASE 1: BiBTeX used to constantly update the references
%   (while the paper is being written).

%\bibliographystyle{informs2014trsc} % outcomment this and next line in Case 1
%\bibliography{sample} % if more than one, comma separated

% CASE 2: BiBTeX used to generate mypaper.bbl (to be further fine tuned)
%\input{mypaper.bbl} % outcomment this line in Case 2

%If you don't use BiBTex, you can manually itemize references as shown below.

%\bibliographystyle{nonumber}

% \begin{thebibliography}{3}
% \providecommand{\natexlab}[1]{#1}
% \providecommand{\url}[1]{\texttt{#1}}
% \providecommand{\urlprefix}{URL }

% \end{thebibliography}

%%%%%%%%%%%%%%%%%
\end{document}